\documentclass[10pt,twocolumn,letterpaper]{article}

\usepackage[pagenumbers]{cvpr}

\usepackage{iftex}
\ifXeTeX
  \usepackage[OT1]{fontenc}
\fi

\usepackage{cuted}

\usepackage{multirow}
\usepackage{tabularx}
\usepackage{colortbl}
\definecolor{xaqamethodbg}{HTML}{E7EEF5}
\definecolor{xaqaevalbg}{HTML}{F3E9EC}
\usepackage{tikz}
\usepackage[skins,breakable]{tcolorbox}

\usepackage{makecell}
\usepackage{pifont}

\newtcolorbox{suppqabox}{
    enhanced jigsaw,
    breakable,
    width=\linewidth,
    colback=white,
    colframe=black!30,
    boxrule=0.45pt,
    sharp corners,
    boxsep=0pt,
    left=4pt,
    right=4pt,
    top=4pt,
    bottom=4pt,
    pad before break=0pt,
    before skip=9pt,
    after skip=12pt,
    fontupper=\small,
    title after break={QA example (continued)},
    fonttitle=\sffamily\footnotesize\itshape,
    colbacktitle=white,
    coltitle=black!65,
    titlerule=0pt,
    toptitle=2pt,
    bottomtitle=1pt,
    lines before break=0
}
\newcommand{\suppturn}[3][10000]{%
    \par\noindent
    \begingroup
    \setlength{\fboxsep}{1.5pt}%
    \colorbox{xaqamethodbg}{\parbox{\dimexpr\linewidth-2\fboxsep\relax}{%
        \sffamily\footnotesize\bfseries\color{black!75}#2}}%
    \endgroup
    \par\penalty#1\relax\vspace{2pt}\noindent #3\par
}

\newcommand{\suppjsonfield}[1]{#1\space\ignorespaces}
\newcommand{\suppjson}[1]{%
    \begingroup
    \clubpenalty=0
    \widowpenalty=0
    \{\ignorespaces #1\unskip\}\par
    \endgroup
}
\newcommand{\suppvideo}{\texttt{\textless video\textgreater}\vadjust{\nobreak}}
\newcommand{\suppimage}{\texttt{\textless image\textgreater}\vadjust{\nobreak}}
\newcommand{\suppqa}[3]{%
    \par\begingroup
    \ifdefined\nolinenumbers\nolinenumbers\fi
    \begin{suppqabox}
        \small\raggedright
        \clubpenalty=10000
        \widowpenalty=10000
        \suppturn{System}{#1}
        \vspace{2pt}
        \suppturn[0]{User}{#2}
        \par\penalty0\vspace{2pt}
        \begingroup
        \clubpenalty=0
        \widowpenalty=0
        \suppturn{Assistant}{#3}
        \endgroup
    \end{suppqabox}%
    \par\endgroup
}

\definecolor{cvprblue}{rgb}{0.21,0.49,0.74}
\usepackage[pagebackref,breaklinks,colorlinks,allcolors=cvprblue]{hyperref}

\def\confName{CVPR}
\def\confYear{2027}

\title{X-AQA: Unifying Action Quality Assessment with Multimodal \\Large Language Models and Multi-Domain Learning}

\author{%
  Kaili Zheng\textsuperscript{1}\quad
  Wei Ding\textsuperscript{2}\quad
  Yuchen Ding\textsuperscript{3}\quad
  Kaiwen Wang\textsuperscript{1}\quad
  Xun Zhu\textsuperscript{1}\\
  Chenyi Guo\textsuperscript{1}\thanks{Corresponding authors:\newline
    Chenyi Guo (\href{mailto:guochy@mail.tsinghua.edu.cn}{\nolinkurl{guochy@mail.tsinghua.edu.cn}})\newline
    Yudong Zhang (\href{mailto:zhangyd16@mails.tsinghua.edu.cn}{\nolinkurl{zhangyd16@mails.tsinghua.edu.cn}})\newline
    Ji Wu (\href{mailto:wuji_ee@mail.tsinghua.edu.cn}{\nolinkurl{wuji_ee@mail.tsinghua.edu.cn}})}\quad
  Yudong Zhang\textsuperscript{1}\footnotemark[1]\quad
  Ji Wu\textsuperscript{1,4,5}\footnotemark[1]\\[3pt]
  \textsuperscript{1}Department of Electronic Engineering, Tsinghua University\\
  \textsuperscript{2}Carnegie Mellon University\quad
  \textsuperscript{3}Baidu, Inc.\quad
  \textsuperscript{4}College of AI, Tsinghua University\\
  \textsuperscript{5}Beijing National Research Center for Information Science and Technology
}

\begin{document}
\maketitle

\begingroup
\preCutedStrip{\vskip-\glueexpr\stripsep+12pt+.5em\relax}
\begin{strip}
  \centering
  \includegraphics[width=\linewidth]{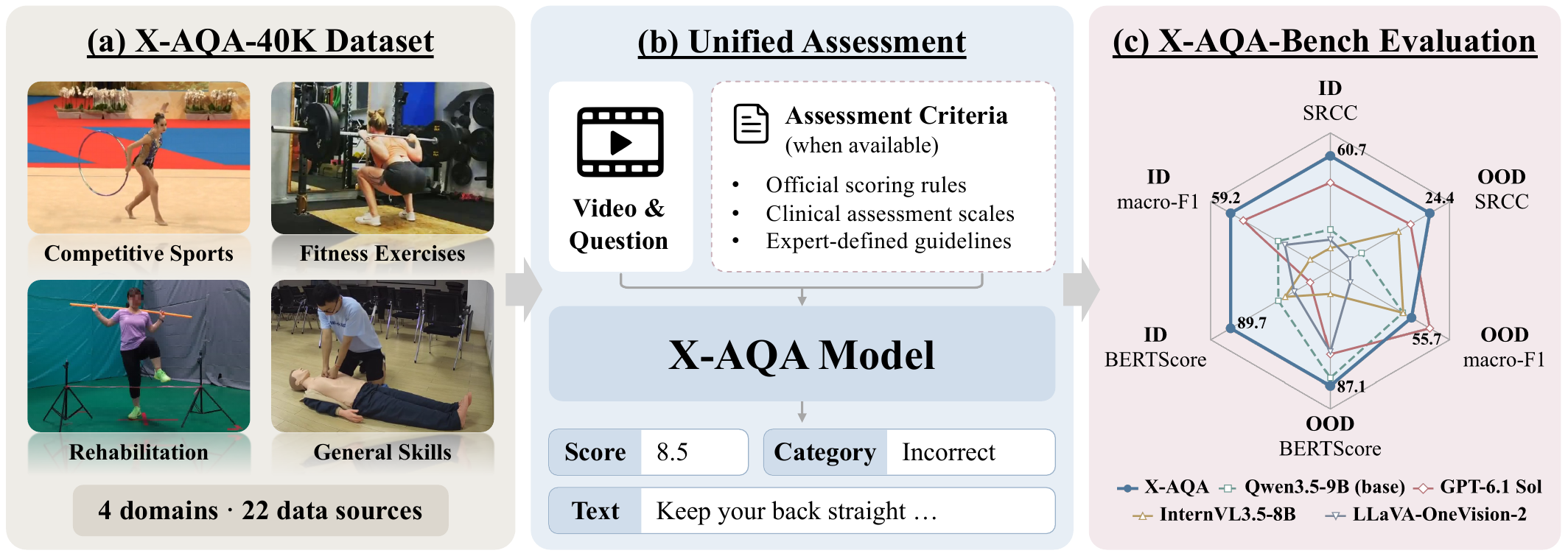}
  \captionsetup{hypcap=false}
  \captionof{figure}{Overview of X-AQA. (a) We curate the X-AQA-40K dataset from 22 data sources across four domains to train a unified AQA model. (b) Through multi-domain learing, we obtain X-AQA, a single MLLM that produces scores, categories, or textual assessments, with assessment criteria provided when available. (c) X-AQA achieves strong performance across assessment tasks on X-AQA-Bench.}
  \captionsetup{hypcap=true}
  \label{fig:teaser}
\end{strip}
\endgroup

\begin{abstract}
Action Quality Assessment (AQA) aims to evaluate the quality of human actions, yet existing methods are predominantly developed and evaluated on individual datasets with different task formulations and assessment criteria.
We introduce \textbf{X-AQA}, the first unified training and evaluation framework for general-purpose AQA with multimodal large language models (MLLMs).
We construct \textbf{X-AQA-40K}, a training corpus comprising 40,163 samples from 22 data sources across four domains: competitive sports, fitness exercises, rehabilitation, and general skills.
We convert heterogeneous annotations into a unified question-answering format, incorporating dataset-specific annotation guidelines and official scoring rules when available.
Using this corpus, we fine-tune Qwen3.5-9B to develop a single MLLM capable of handling diverse AQA tasks.
We further establish \textbf{X-AQA-Bench}, which evaluates both in-distribution (ID) performance on datasets represented during training and out-of-distribution (OOD) generalization on four held-out datasets.
The resulting model outperforms all evaluated open-source MLLMs on X-AQA-Bench and achieves competitive performance relative to proprietary MLLMs.
Ablation studies further demonstrate that multi-domain joint training and explicitly providing assessment criteria in prompts both improve OOD generalization.
X-AQA establishes a unified foundation for advancing general-purpose AQA across diverse real-world domains.
Data and models are released at \href{https://huggingface.co/collections/Kelly0510/x-aqa}{here}.
\end{abstract}

\section{Introduction}
\label{sec:intro}

Action Quality Assessment (AQA) aims to evaluate how well human actions are performed~\cite{pirsiavash2014assessing}, with applications spanning competitive sports~\cite{xu2022finediving}, fitness exercises~\cite{fitnessaqa}, rehabilitation~\cite{li2024finerehab}, and general skills~\cite{doughty2018whos}.
Unlike action recognition, which focuses on what action is performed, AQA evaluates how well the action is performed according to task-specific objectives and assessment criteria.
AQA tasks take diverse forms, including performance scoring, execution error detection, relative skill assessment, and textual feedback generation~\cite{mtlaqa,expertaf}.
Assessment criteria also vary across domains, including official scoring rules in competitive sports, standardized clinical assessment scales in rehabilitation, and expert-defined guidelines for identifying execution errors.
Due to this diversity, existing AQA methods are typically developed for individual datasets, with architectures and learning objectives tailored to their assessment tasks and supervision.

To move beyond dataset-specific AQA, recent studies have explored cross-action learning. 
Specifically, Parmar and Morris~\cite{aqa7} train a shared C3D-LSTM model across multiple action categories, while MoAKE~\cite{xu2026moake} introduces a mixture-of-experts architecture to capture diverse action patterns and adaptively aggregate action knowledge.
Although these methods demonstrate the benefits of cross-action learning, they primarily focus on numerical score prediction. 
AQA, however, encompasses a broader range of assessment tasks, including numerical scoring, categorical judgments, and textual feedback. 
Multimodal large language models (MLLMs) offer a promising foundation for unifying these tasks through their multimodal understanding and instruction-following capabilities~\cite{lin2024videollava,bai2025qwen3}.
Recent studies have examined MLLMs for AQA, identifying challenges in fine-grained execution understanding and quality judgment~\cite{freitas2026judge,zheng2026fitaqa}.
However, how to effectively train MLLMs using heterogeneous AQA supervision and evaluate their generalization to unseen datasets remains underexplored.

In this paper, we introduce X-AQA, the first unified training and evaluation framework for general-purpose AQA with MLLMs, as illustrated in \cref{fig:teaser}.
We construct X-AQA-40K, a training corpus comprising 40,163 samples from 22 datasets across four domains: competitive sports, fitness exercises, rehabilitation, and general skills. 
We unify heterogeneous annotations into question-answer pairs covering numerical scoring, categorical judgments, and textual feedback, while preserving the assessment objectives and label semantics of each source. 
When available, dataset-specific annotation guidelines and official scoring rules are incorporated into the prompts to make the assessment criteria explicit. 
Using this corpus, we perform multi-domain joint training to develop a single MLLM capable of addressing diverse AQA tasks.
We further establish X-AQA-Bench to systematically evaluate AQA performance and generalization. 
It includes in-distribution (ID) evaluation on held-out samples from datasets represented during training and out-of-distribution (OOD) evaluation on four held-out source datasets. 

We fine-tune Qwen3.5-9B~\cite{qwen35} on X-AQA-40K to obtain X-AQA, a unified AQA model. On the ID split of X-AQA-Bench, X-AQA surpasses all evaluated open-source and proprietary MLLMs on all four metrics. In particular, X-AQA achieves an SRCC of 0.607, compared with 0.465 for GPT-6.1 Sol~\cite{openai2026gpt61sol}. On the OOD split, X-AQA outperforms all evaluated open-source MLLMs on all four metrics and achieves competitive performance relative to proprietary MLLMs. For textual feedback, X-AQA achieves a BERTScore of 0.871, compared with 0.862 for Gemini 3.8 Flash~\cite{google2026gemini38flash}. Ablation studies further demonstrate the benefits of multi-domain learning and explicit assessment criteria for generalization to unseen datasets. On the OOD split, multi-domain learning raises the SRCC from 0.107 to 0.244, while providing assessment criteria increases macro-F1 from 0.470 to 0.557. Overall, X-AQA can address diverse AQA tasks and generalize to unseen datasets.

Our main contributions are summarized as follows:
\begin{itemize}
\item We construct X-AQA-40K, a training corpus comprising samples from 22 datasets across four domains. It brings heterogeneous assessment tasks and available assessment criteria into a shared question-answering format.
\item We establish X-AQA-Bench, a unified benchmark spanning numerical, categorical, and textual assessment, with ID and OOD splits for evaluating AQA performance and generalization.
\item We develop X-AQA, a unified MLLM for diverse AQA tasks that outperforms all evaluated open-source MLLMs across all benchmark metrics and achieves competitive performance relative to proprietary MLLMs. Ablation studies show that multi-domain learning and explicit assessment criteria improve OOD generalization.
\end{itemize}

\section{Related Work}
\label{sec:relate}

\begin{figure*}[!t]
    \centering
    \vbox{%
        \hsize=\textwidth
        \centering
        \offinterlineskip
        \begingroup
\color{black}
\definecolor{xaqatrainsports}{HTML}{BD7D91}
\definecolor{xaqatrainfitness}{HTML}{7499BC}
\definecolor{xaqatrainrehab}{HTML}{75A389}
\definecolor{xaqatraingeneral}{HTML}{C4A064}
\newcommand{\xaqatrainsector}[5]{%
  \path[fill=#1,draw=white,line width=.3pt]
    (#2:#4) -- (#2:#5)
    arc[start angle=#2,end angle=#3,radius=#5pt] -- (#3:#4)
    arc[start angle=#3,end angle=#2,radius=#4pt] -- cycle;
}
\newcommand{\xaqatrainheading}[5]{%
  \path[fill=#2!25!white] (152,{#1-5.5}) rectangle (496,{#1+5.5});
  \node[anchor=west,font=\fontsize{8.5}{10}\selectfont\bfseries]
    at (157,#1) {#3};
  \node[anchor=east,font=\fontsize{8.5}{10}\selectfont]
    at (491,#1) {#4\enspace(#5\%)};
}
\newcommand{\xaqatrainentry}[5]{%
  \path[fill=#3]
    (#1,{#2-2.4}) rectangle ({#1+4.8},{#2+2.4});
  \node[anchor=west] at ({#1+7.5},#2) {#4\,(#5)};
}
\begin{tikzpicture}[x=1pt,y=1pt,
  every node/.style={inner sep=0pt,outer sep=0pt,text=black,
    font=\fontsize{8}{9.5}\selectfont}]
  \path[use as bounding box] (0,0) rectangle (496,147);
  \begin{scope}[shift={(69,73.5)}]
    \xaqatrainsector{xaqatrainsports!65!white}{90.000000}{-34.628140}{34}{50}
    \xaqatrainsector{xaqatrainsports!100!white}{90.000000}{72.073052}{51}{67}
    \xaqatrainsector{xaqatrainsports!91!white}{72.073052}{54.146105}{51}{67}
    \xaqatrainsector{xaqatrainsports!82!white}{54.146105}{36.219157}{51}{67}
    \xaqatrainsector{xaqatrainsports!73!white}{36.219157}{18.471479}{51}{67}
    \xaqatrainsector{xaqatrainsports!64!white}{18.471479}{3.565222}{51}{67}
    \xaqatrainsector{xaqatrainsports!55!white}{3.565222}{-10.767373}{51}{67}
    \xaqatrainsector{xaqatrainsports!46!white}{-10.767373}{-20.259692}{51}{67}
    \xaqatrainsector{xaqatrainsports!37!white}{-20.259692}{-27.457361}{51}{67}
    \xaqatrainsector{xaqatrainsports!28!white}{-27.457361}{-34.628140}{51}{67}
    \xaqatrainsector{xaqatrainfitness!65!white}{-34.628140}{-150.938177}{34}{50}
    \xaqatrainsector{xaqatrainfitness!100!white}{-34.628140}{-70.482036}{51}{67}
    \xaqatrainsector{xaqatrainfitness!82!white}{-70.482036}{-94.109753}{51}{67}
    \xaqatrainsector{xaqatrainfitness!64!white}{-94.109753}{-115.084282}{51}{67}
    \xaqatrainsector{xaqatrainfitness!46!white}{-115.084282}{-133.011229}{51}{67}
    \xaqatrainsector{xaqatrainfitness!28!white}{-133.011229}{-150.938177}{51}{67}
    \xaqatrainsector{xaqatrainrehab!65!white}{-150.938177}{-184.362971}{34}{50}
    \xaqatrainsector{xaqatrainrehab!100!white}{-150.938177}{-163.944177}{51}{67}
    \xaqatrainsector{xaqatrainrehab!82!white}{-163.944177}{-170.460623}{51}{67}
    \xaqatrainsector{xaqatrainrehab!64!white}{-170.460623}{-175.838707}{51}{67}
    \xaqatrainsector{xaqatrainrehab!46!white}{-175.838707}{-180.678983}{51}{67}
    \xaqatrainsector{xaqatrainrehab!28!white}{-180.678983}{-184.362971}{51}{67}
    \xaqatrainsector{xaqatraingeneral!65!white}{-184.362971}{-270.000000}{34}{50}
    \xaqatrainsector{xaqatraingeneral!100!white}{-184.362971}{-210.410826}{51}{67}
    \xaqatrainsector{xaqatraingeneral!82!white}{-210.410826}{-228.337774}{51}{67}
    \xaqatrainsector{xaqatraingeneral!64!white}{-228.337774}{-246.264721}{51}{67}
    \xaqatrainsector{xaqatraingeneral!46!white}{-246.264721}{-264.191669}{51}{67}
    \xaqatrainsector{xaqatraingeneral!28!white}{-264.191669}{-270.000000}{51}{67}
    \node[anchor=center] at (0,0) {\shortstack[c]{%
      {\fontsize{10.5}{12.5}\selectfont\bfseries X-AQA-40K}\\[1pt]
      {\fontsize{12}{14}\selectfont\bfseries 40,163}\\[-2pt]
      training samples}};
  \end{scope}
  \xaqatrainheading{140.0}{xaqatrainsports}{Competitive Sports}{13,904}{34.6}
  \xaqatrainentry{153}{127.5}{xaqatrainsports!100!white}{FineFS~\cite{finefs}}{2,000}
  \xaqatrainentry{261}{127.5}{xaqatrainsports!91!white}{CricketVision~\cite{ujaqa}}{2,000}
  \xaqatrainentry{389}{127.5}{xaqatrainsports!82!white}{ExpertAF~\cite{expertaf}}{2,000}
  \xaqatrainentry{153}{118.0}{xaqatrainsports!73!white}{Ego-Exo4D~\cite{egoexo4d}}{1,980}
  \xaqatrainentry{261}{118.0}{xaqatrainsports!64!white}{MultiSenseBadminton~\cite{multisensebadminton}}{1,663}
  \xaqatrainentry{389}{118.0}{xaqatrainsports!55!white}{CoT-AFA~\cite{cotafa}}{1,599}
  \xaqatrainentry{153}{108.5}{xaqatrainsports!46!white}{MTL-AQA~\cite{mtlaqa}}{1,059}
  \xaqatrainentry{261}{108.5}{xaqatrainsports!37!white}{AQA-7~\cite{aqa7}}{803}
  \xaqatrainentry{389}{108.5}{xaqatrainsports!28!white}{RG~\cite{rg}}{800}
  \xaqatrainheading{96.0}{xaqatrainfitness}{Fitness Exercises}{12,976}{32.3}
  \xaqatrainentry{153}{83.5}{xaqatrainfitness!100!white}{QEVD~\cite{qevd}}{4,000}
  \xaqatrainentry{261}{83.5}{xaqatrainfitness!82!white}{EgoExo-Fitness~\cite{egoexofitness}}{2,636}
  \xaqatrainentry{389}{83.5}{xaqatrainfitness!64!white}{CoT-AFA~\cite{cotafa}}{2,340}
  \xaqatrainentry{153}{74.0}{xaqatrainfitness!46!white}{Fitness-AQA~\cite{fitnessaqa}}{2,000}
  \xaqatrainentry{261}{74.0}{xaqatrainfitness!28!white}{FLAG3D~\cite{flag3d,flag3dpp}}{2,000}
  \xaqatrainheading{61.5}{xaqatrainrehab}{Rehabilitation}{3,729}{9.3}
  \xaqatrainentry{153}{49.0}{xaqatrainrehab!100!white}{LLM-FMS~\cite{llmfms}}{1,451}
  \xaqatrainentry{261}{49.0}{xaqatrainrehab!82!white}{REHAB24-6~\cite{rehab24paper,rehab24}}{727}
  \xaqatrainentry{389}{49.0}{xaqatrainrehab!64!white}{ExeCheck~\cite{execheck}}{600}
  \xaqatrainentry{153}{39.5}{xaqatrainrehab!46!white}{KERAAL~\cite{keraal}}{540}
  \xaqatrainentry{261}{39.5}{xaqatrainrehab!28!white}{UpperLimb~\cite{upperlimb}}{411}
  \xaqatrainheading{27.0}{xaqatraingeneral}{General Skills}{9,554}{23.8}
  \xaqatrainentry{153}{14.5}{xaqatraingeneral!100!white}{LASANA~\cite{lasana}}{2,906}
  \xaqatrainentry{261}{14.5}{xaqatraingeneral!82!white}{CaptainCook4D~\cite{captaincook4d}}{2,000}
  \xaqatrainentry{389}{14.5}{xaqatraingeneral!64!white}{Ego-Exo4D~\cite{egoexo4d}}{2,000}
  \xaqatrainentry{153}{5.0}{xaqatraingeneral!46!white}{AIxSuture~\cite{hoffmann2024aixsuture}}{2,000}
  \xaqatrainentry{261}{5.0}{xaqatraingeneral!28!white}{CPR-Coach~\cite{cprcoach}}{648}
\end{tikzpicture}
\endgroup\par
        \vskip7pt
        \begingroup
\color{black}
\definecolor{xaqatrainsports}{HTML}{BD7D91}
\definecolor{xaqatrainfitness}{HTML}{7499BC}
\definecolor{xaqatrainrehab}{HTML}{75A389}
\definecolor{xaqatraingeneral}{HTML}{C4A064}
\begin{tikzpicture}[x=1pt,y=1pt,every node/.style={inner sep=0pt,outer sep=0pt,text=black}]
\path[use as bounding box] (0,0) rectangle (496,47);
\node[anchor=center,rotate=0,font=\fontsize{10}{12}\selectfont\bfseries] at (248.0000,41) {X-AQA-Bench {\mdseries (7,101 test samples)}};
\node[anchor=west,font=\fontsize{9}{11}\selectfont\bfseries] (benchmarkood) at (0,6.5) {OOD};
\node[anchor=east,font=\fontsize{9}{11}\selectfont\bfseries] at (benchmarkood.east |- 0,25.5) {ID};
\path[fill=xaqatrainsports!100!white,draw=white,line width=.3pt] (25.5000,19) rectangle (51.8045,32);
\node[anchor=center,rotate=0,font=\fontsize{8}{10}\selectfont] at (38.6523,25.5) {300};
\path[fill=xaqatrainsports!91!white,draw=white,line width=.3pt] (51.8045,19) rectangle (55.3118,32);
\path[fill=xaqatrainsports!82!white,draw=white,line width=.3pt] (55.3118,19) rectangle (60.8357,32);
\path[fill=xaqatrainsports!73!white,draw=white,line width=.3pt] (60.8357,19) rectangle (84.6851,32);
\node[anchor=center,rotate=0,font=\fontsize{8}{10}\selectfont] at (72.7604,25.5) {272};
\path[fill=xaqatrainsports!64!white,draw=white,line width=.3pt] (84.6851,19) rectangle (85.2112,32);
\path[fill=xaqatrainsports!55!white,draw=white,line width=.3pt] (85.2112,19) rectangle (115.8998,32);
\node[anchor=center,rotate=0,font=\fontsize{8}{10}\selectfont] at (100.5555,25.5) {350};
\path[fill=xaqatrainsports!46!white,draw=white,line width=.3pt] (115.8998,19) rectangle (145.6239,32);
\node[anchor=center,rotate=0,font=\fontsize{8}{10}\selectfont] at (130.7619,25.5) {339};
\path[fill=xaqatrainsports!37!white,draw=white,line width=.3pt] (145.6239,19) rectangle (171.9284,32);
\node[anchor=center,rotate=0,font=\fontsize{8}{10}\selectfont] at (158.7762,25.5) {300};
\path[fill=xaqatrainsports!28!white,draw=white,line width=.3pt] (171.9284,19) rectangle (189.4648,32);
\node[anchor=center,rotate=0,font=\fontsize{8}{10}\selectfont] at (180.6966,25.5) {200};
\path[fill=xaqatrainfitness!100!white,draw=white,line width=.3pt] (189.4648,19) rectangle (215.7693,32);
\node[anchor=center,rotate=0,font=\fontsize{8}{10}\selectfont] at (202.6170,25.5) {300};
\path[fill=xaqatrainfitness!82!white,draw=white,line width=.3pt] (215.7693,19) rectangle (263.4681,32);
\node[anchor=center,rotate=0,font=\fontsize{8}{10}\selectfont] at (239.6187,25.5) {544};
\path[fill=xaqatrainfitness!64!white,draw=white,line width=.3pt] (263.4681,19) rectangle (316.0772,32);
\node[anchor=center,rotate=0,font=\fontsize{8}{10}\selectfont] at (289.7726,25.5) {600};
\path[fill=xaqatrainfitness!46!white,draw=white,line width=.3pt] (316.0772,19) rectangle (342.3817,32);
\node[anchor=center,rotate=0,font=\fontsize{8}{10}\selectfont] at (329.2294,25.5) {300};
\path[fill=xaqatrainfitness!28!white,draw=white,line width=.3pt] (342.3817,19) rectangle (368.6862,32);
\node[anchor=center,rotate=0,font=\fontsize{8}{10}\selectfont] at (355.5339,25.5) {300};
\path[fill=xaqatrainrehab!100!white,draw=white,line width=.3pt] (368.6862,19) rectangle (394.9907,32);
\node[anchor=center,rotate=0,font=\fontsize{8}{10}\selectfont] at (381.8384,25.5) {300};
\path[fill=xaqatrainrehab!82!white,draw=white,line width=.3pt] (394.9907,19) rectangle (397.0074,32);
\path[fill=xaqatrainrehab!64!white,draw=white,line width=.3pt] (397.0074,19) rectangle (398.7610,32);
\path[fill=xaqatrainrehab!28!white,draw=white,line width=.3pt] (398.7610,19) rectangle (405.7755,32);
\node[anchor=center,rotate=0,font=\fontsize{8}{10}\selectfont] at (402.2683,25.5) {80};
\path[fill=xaqatraingeneral!100!white,draw=white,line width=.3pt] (405.7755,19) rectangle (440.8482,32);
\node[anchor=center,rotate=0,font=\fontsize{8}{10}\selectfont] at (423.3119,25.5) {400};
\path[fill=xaqatraingeneral!82!white,draw=white,line width=.3pt] (440.8482,19) rectangle (449.2657,32);
\node[anchor=center,rotate=0,font=\fontsize{8}{10}\selectfont] at (445.0569,25.5) {96};
\path[fill=xaqatraingeneral!64!white,draw=white,line width=.3pt] (449.2657,19) rectangle (465.6621,32);
\node[anchor=center,rotate=0,font=\fontsize{8}{10}\selectfont] at (457.4639,25.5) {187};
\path[fill=xaqatraingeneral!46!white,draw=white,line width=.3pt] (465.6621,19) rectangle (469.6955,32);
\path[fill=xaqatraingeneral!28!white,draw=white,line width=.3pt] (469.6955,19) rectangle (496.0000,32);
\node[anchor=center,rotate=0,font=\fontsize{8}{10}\selectfont] at (482.8477,25.5) {300};
\path[fill=xaqatrainsports!25!white,draw=white,line width=.3pt] (25.5000,0) rectangle (215.3271,13);
\node[anchor=center,rotate=0,font=\fontsize{8}{10}\selectfont] at (120.4135,6.5) {T3Set~\cite{t3set} (700)};
\path[fill=xaqatrainfitness!25!white,draw=white,line width=.3pt] (215.3271,0) rectangle (309.6983,13);
\node[anchor=center,rotate=0,font=\fontsize{8}{10}\selectfont] at (262.5127,6.5) {Fitness-7~\cite{fitness7} (348)};
\path[fill=xaqatrainrehab!25!white,draw=white,line width=.3pt] (309.6983,0) rectangle (404.0695,13);
\node[anchor=center,rotate=0,font=\fontsize{8}{10}\selectfont] at (356.8839,6.5) {KIMORE~\cite{capecci2018instrumental} (348)};
\path[fill=xaqatraingeneral!25!white,draw=white,line width=.3pt] (404.0695,0) rectangle (496.0000,13);
\node[anchor=center,rotate=0,font=\fontsize{8}{10}\selectfont] at (450.0347,6.5) {JIGSAWS~\cite{jigsaws} (339)};
\end{tikzpicture}
\endgroup\par
    }
    \caption{Composition of X-AQA-40K and X-AQA-Bench. X-AQA-40K comprises 40,163 training samples from 22 data sources across four domains, with Ego-Exo4D and CoT-AFA each spanning two domains. X-AQA-Bench contains 7,101 test samples, including 5,366 ID samples from datasets represented during training and 1,735 OOD samples from four held-out datasets. RG denotes Rhythmic Gymnastics.}
    \label{fig:training_data}
\end{figure*}
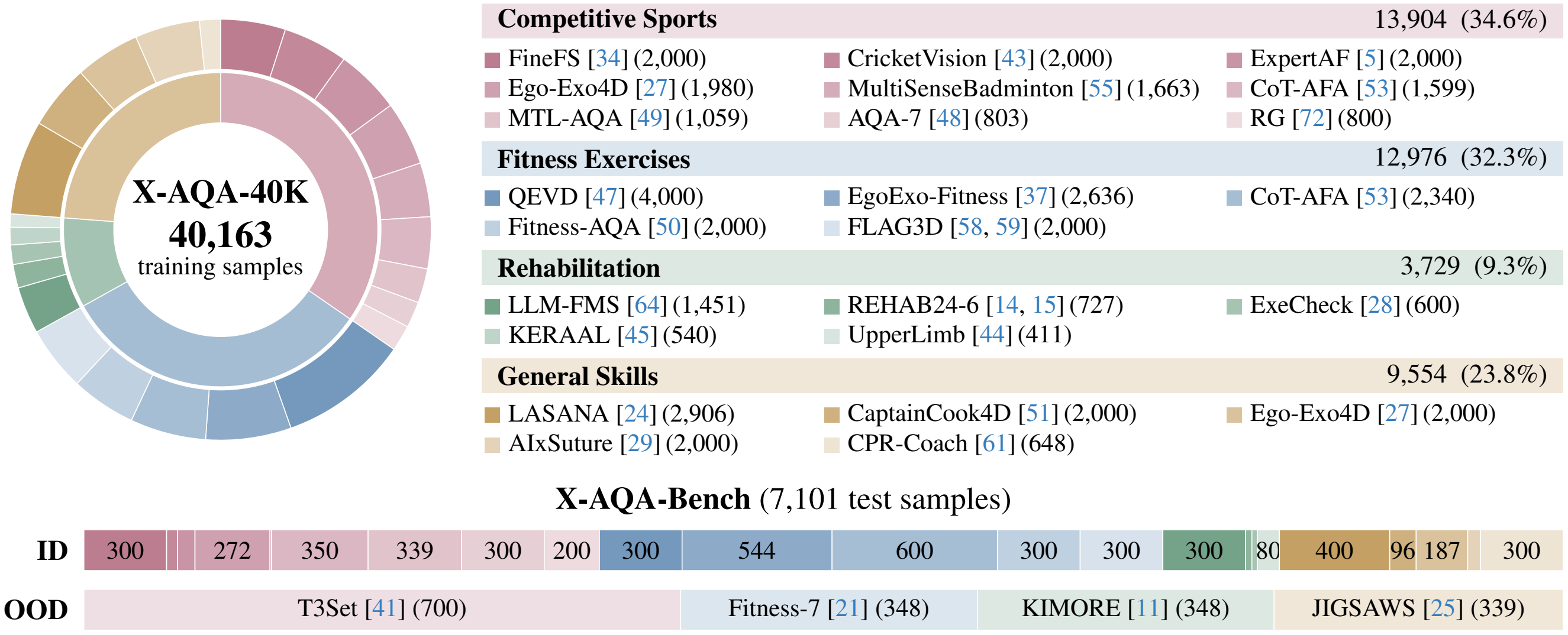

\subsection{Action Quality Assessment}
AQA has been studied across datasets spanning diverse domains, with different assessment objectives and annotation schemes.
Numerical scoring is widely studied in competitive sports, with methods employing direct score regression or reference-based contrastive regression~\cite{yu2021core}.
Beyond the choice of regression strategy, some methods explicitly model how an overall performance score is composed.
For example, FineGrade~\cite{li2026finegrade} predicts difficulty and execution scores separately and enforces their additive relationship according to official gymnastics scoring rules.
Numerical assessment is also used in rehabilitation, where KIMORE~\cite{capecci2019kimore,capecci2018hsmm,capecci2018instrumental} provides clinician-assigned scores for therapeutic exercises.
Beyond numerical scoring, categorical assessment addresses tasks such as execution error identification and proficiency classification.
For instance, FLEX~\cite{yin2025flex} provides annotations of exercise execution errors, while Ego-Exo4D~\cite{egoexo4d} includes proficiency annotations for diverse physical skills.
Textual assessment provides descriptive feedback on action execution.
MTL-AQA explores commentary generation alongside score prediction, while FLEX and Ego-Exo4D incorporate corrective feedback and expert commentary, respectively.
Overall, existing AQA methods are typically developed and evaluated within the task settings defined by individual datasets.

\subsection{Cross-Action Learning for AQA}
Beyond dataset-specific AQA, several studies have explored sharing assessment knowledge across different action categories.
For example, Parmar and Morris investigate cross-action knowledge transfer by jointly training a shared C3D-LSTM model on multiple actions.
MoAKE introduces a mixture-of-experts architecture to capture complementary action knowledge and adaptively aggregate it across actions, demonstrating generalization to unseen actions under zero- and few-shot settings.
However, both approaches focus on numerical score prediction.
CrossTrainer~\cite{ashutosh2025crosstrainer} extends cross-action learning to language-based assessment by learning transferable skill attributes from expert commentary, but focuses on proficiency estimation and actionable feedback generation across sports.
A unified model that jointly addresses numerical scoring, categorical judgments, and textual feedback across diverse actions remains largely unexplored.

\subsection{MLLMs for AQA}
MLLMs have been explored for action assessment tasks that involve quality judgments and natural-language feedback.
ExpertAF develops a multimodal model that generates expert commentary and corrective demonstrations from action videos and body poses.
ProfVLM~\cite{profvlm2026} formulates multi-view proficiency assessment as conditional language generation, jointly predicting proficiency levels and expert-like commentary.
For interactive fitness coaching, Panchal \etal~\cite{qevd} introduce the QEVD benchmark and a streaming model that provides appropriate feedback during ongoing exercises.
Beyond developing task-specific models, recent studies have systematically evaluated the AQA capabilities of general-purpose MLLMs.
Freitas \etal investigate model performance across AQA domains, tasks, and prompting strategies.
FitAQA evaluates fine-grained visual perception, correctness judgment, and temporal error localization.
However, existing research has yet to establish a unified framework for training MLLMs on heterogeneous AQA supervision and evaluating their generalization across diverse AQA tasks and datasets.

\section{Method}
\label{sec:method}

We construct X-AQA-40K by integrating heterogeneous datasets into a shared question-answering format and incorporating assessment criteria when available.
We then jointly train a single MLLM on this corpus and establish X-AQA-Bench to evaluate performance and generalization.

\subsection{X-AQA-40K Dataset}
\label{sec:data_curation}

\paragraph{Data Composition.}
We construct X-AQA-40K from 22 existing data sources spanning four domains, competitive sports, fitness exercises, rehabilitation, and general skills, as illustrated in \cref{fig:training_data} (top).
These sources cover diverse AQA tasks, such as diving execution scoring in MTL-AQA, exercise error detection in Fitness-AQA, rehabilitation correctness assessment in REHAB24-6~\cite{rehab24paper}, suturing skill assessment in AIxSuture~\cite{hoffmann2024aixsuture}, and coaching feedback generation in QEVD.
To balance training data across sources and tasks, we cap each dataset-specific task at 2,000 training samples.
For Ego-Exo4D and CoT-AFA~\cite{cotafa}, which span multiple domains, the cap is applied separately to each task within each domain.
Datasets with multiple assessment tasks contribute separate training subsets.
Examples include execution question answering and live coaching feedback in QEVD, technical-point assessment and overall feedback in EgoExo-Fitness~\cite{egoexofitness}, form assessment and explanatory feedback in CoT-AFA, and skill scoring and error detection in LASANA~\cite{lasana}.
For FineFS~\cite{finefs} and Rhythmic Gymnastics~\cite{rg}, we mask visible broadcast scoreboards to prevent direct leakage of ground-truth scores.
The resulting X-AQA-40K corpus contains 40,163 question-answer pairs across 29 task-specific subsets, covering 19,614 unique media files.
Detailed task-level dataset statistics and preprocessing procedures for each source dataset are provided in the supplementary material.

\paragraph{Unified Question-Answering Formulation.}
We convert heterogeneous AQA annotations into a unified question-answering format.
Each training sample consists of a visual input, a task-specific assessment instruction, and a reference answer derived from the original annotations.
Numerical scores and categorical judgments are represented using task-specific JSON fields, while textual feedback is expressed in natural language.
For tasks involving multiple assessment aspects, a single answer can contain multiple fields, as in MTL-AQA, which combines execution scoring and commentary.
This formulation enables a single MLLM to learn diverse assessment tasks through a shared interface while preserving their task-specific output semantics.

\paragraph{Assessment Criteria Integration.}
Assessing action quality requires task-specific criteria that define what constitutes good execution.
To provide explicit guidance across diverse AQA tasks, we incorporate assessment criteria into the assessment prompts when available, drawing from official sports scoring rules, clinical assessment scales, or expert-defined annotation guidelines.
For competitive sports, we draw on official scoring rules, including FINA diving rules~\cite{worldaquaticsdivingrules} for MTL-AQA, ISU figure skating rules~\cite{isufigureskatingrules} for FineFS, and FIG rhythmic gymnastics rules~\cite{figrules} for Rhythmic Gymnastics, using the rule editions applicable to the corresponding competitions.
For datasets with established assessment rubrics, we incorporate their scoring dimensions, rating scales, and error definitions, such as OSATS in AIxSuture, FMS criteria in LLM-FMS~\cite{llmfms}, and skill assessment guidelines in LASANA.
For fitness exercises, we formulate assessment criteria from the annotated form errors and technical points in Fitness-AQA and EgoExo-Fitness.
For REHAB24-6, we derive observable correctness criteria from the provided rehabilitation exercise descriptions.
These criteria describe general assessment standards independently of each sample's reference answer and accompany the task instructions and output specifications.
We follow this approach during both training and evaluation.

\subsection{Multi-Domain Joint Training}
\label{sec:joint_training}

We jointly fine-tune Qwen3.5-9B on X-AQA-40K to learn diverse AQA tasks across all four domains.
Given a training sample consisting of visual input $V_i$, assessment instruction $Q_i$, and reference answer $A_i$, we optimize the autoregressive language modeling objective:
\begin{equation}
    \mathcal{L}(\theta) = -\frac{1}{\sum_i |A_i|}
    \sum_i w_i \sum_{t=1}^{|A_i|} \log P_{\theta}(a_{i,t}\mid V_i,Q_i,a_{i,<t})
    \label{eq:training_loss}
\end{equation}
where $\theta$ denotes the trainable model parameters, $|A_i|$ is the number of tokens in the reference answer, $a_{i,t}$ is its $t$-th token, and $a_{i,<t}$ is the preceding answer prefix.
The sample-specific loss weight is denoted by $w_i$.
Similar to RAS~\cite{cao2025refer}, we assign higher loss weights to underrepresented classes.
For categorical AQA tasks, we compute these weights from label frequencies within task-specific groups, such as individual exercises or technical points.
The weights are normalized within each group and capped at 20 to limit extreme values, balancing classification supervision without altering the training sample distribution.
All assessment tasks are optimized using this shared objective, without task-specific prediction heads.

\begin{table*}[htbp]
\centering
\caption{Comparison with general-purpose MLLMs on X-AQA-Bench. All metrics are higher-is-better ($\uparrow$), and the best result among open-source models in each column is highlighted in bold. All reported values are multiplied by 100. Subscripts indicate the half-widths of 95\% bootstrap confidence intervals.}
\label{tab:benchmark_main}
\small
\begin{tabularx}{\textwidth}{X*{8}{c}}
\toprule
\multirow{3}{*}{\textbf{Method}}
& \multicolumn{4}{c}{\textbf{ID}} & \multicolumn{4}{c}{\textbf{OOD}} \\
\cmidrule(lr){2-5}\cmidrule(lr){6-9}
& \multicolumn{1}{c}{Score} & \multicolumn{2}{c}{Category} & \multicolumn{1}{c}{Text}
& \multicolumn{1}{c}{Score} & \multicolumn{2}{c}{Category} & \multicolumn{1}{c}{Text} \\
\cmidrule(lr){2-2}\cmidrule(lr){3-4}\cmidrule(lr){5-5}\cmidrule(lr){6-6}\cmidrule(lr){7-8}\cmidrule(lr){9-9}
& SRCC$\uparrow$ & macro-F1$\uparrow$ & $\kappa\uparrow$ & BERTScore$\uparrow$
& SRCC$\uparrow$ & macro-F1$\uparrow$ & $\kappa\uparrow$ & BERTScore$\uparrow$ \\
\midrule
\multicolumn{9}{c}{\textit{Baselines}} \\
Chance Level (Random) & -- & 42.7$_{1.8}$ & -1.8$_{3.3}$ & -- & -- & 48.3$_{5.3}$ & -3.4$_{10.4}$ & -- \\
Chance Level (Frequency) & -- & 33.4$_{0.7}$ & 0.0$_{0.0}$ & -- & -- & 33.3$_{2.3}$ & 0.0$_{0.0}$ & -- \\
\midrule
\multicolumn{9}{c}{\textit{Open-Source Models}} \\
Molmo2-8B & 8.8$_{5.5}$ & 39.6$_{1.7}$ & 4.2$_{2.6}$ & 86.5$_{0.1}$ & 15.6$_{6.8}$ & 34.9$_{3.1}$ & 0.6$_{2.4}$ & 86.0$_{0.1}$ \\
VideoLLaMA3-7B & -7.0$_{7.7}$ & 44.3$_{2.0}$ & 4.2$_{3.1}$ & 87.0$_{0.1}$ & 7.7$_{5.2}$ & 34.2$_{2.9}$ & -0.6$_{2.6}$ & 85.6$_{0.1}$ \\
InternVL3.5-8B & 12.1$_{5.2}$ & 41.6$_{1.7}$ & 6.5$_{2.9}$ & 87.3$_{0.1}$ & 16.1$_{8.7}$ & 53.2$_{5.3}$ & 6.3$_{10.5}$ & 83.6$_{0.1}$ \\
Keye-VL-1.5-8B & 13.9$_{6.5}$ & 42.1$_{1.9}$ & 6.7$_{3.0}$ & 87.4$_{0.1}$ & 11.9$_{8.2}$ & 33.4$_{2.6}$ & -1.7$_{2.6}$ & 84.9$_{0.1}$ \\
MiniCPM-V 4.5 & 11.8$_{5.9}$ & 43.4$_{1.9}$ & 6.5$_{3.1}$ & 87.2$_{0.1}$ & 13.4$_{8.3}$ & 39.6$_{4.3}$ & 3.4$_{4.7}$ & 86.1$_{0.1}$ \\
InternVideo3-8B-Instruct & 20.6$_{5.4}$ & 45.0$_{1.8}$ & 8.7$_{3.1}$ & 87.4$_{0.1}$ & 12.7$_{8.1}$ & 50.0$_{5.3}$ & 10.9$_{7.9}$ & 86.5$_{0.1}$ \\
Eagle2.5-8B & 18.8$_{4.9}$ & 45.5$_{1.8}$ & 6.8$_{3.5}$ & 87.1$_{0.1}$ & 19.9$_{7.9}$ & 39.7$_{4.2}$ & 5.2$_{4.0}$ & 85.4$_{0.1}$ \\
LLaVA-OneVision-2 & 16.5$_{6.0}$ & 47.2$_{1.9}$ & 10.7$_{3.4}$ & 86.9$_{0.1}$ & 3.3$_{7.8}$ & 37.3$_{3.9}$ & 1.1$_{4.2}$ & 85.8$_{0.1}$ \\
Qwen2.5-VL-7B-Instruct & 5.4$_{5.2}$ & 43.6$_{1.9}$ & 6.7$_{3.1}$ & 87.1$_{0.1}$ & 20.7$_{7.8}$ & 33.2$_{2.3}$ & -0.6$_{0.9}$ & 85.5$_{0.1}$ \\
Qwen3-VL-8B-Instruct & 22.4$_{5.2}$ & 45.9$_{1.9}$ & 7.0$_{3.4}$ & 87.1$_{0.1}$ & 12.9$_{7.1}$ & 46.1$_{5.1}$ & 4.6$_{7.6}$ & 86.5$_{0.1}$ \\
Qwen3-VL-8B-Thinking & 11.7$_{5.8}$ & 48.1$_{1.9}$ & 12.5$_{3.3}$ & 86.5$_{0.1}$ & 8.9$_{7.6}$ & 52.7$_{5.3}$ & 8.1$_{9.9}$ & 86.2$_{0.1}$ \\
Qwen3.5-9B, Non-Thinking & 21.8$_{5.0}$ & 48.7$_{1.8}$ & 14.1$_{3.3}$ & 87.6$_{0.1}$ & 6.3$_{4.2}$ & 53.1$_{5.3}$ & 6.9$_{10.4}$ & 86.8$_{0.1}$ \\
Qwen3.5-9B, Thinking & 19.3$_{8.2}$ & 50.2$_{1.9}$ & 13.6$_{3.4}$ & 86.7$_{0.1}$ & 14.4$_{8.0}$ & 49.6$_{5.3}$ & 5.2$_{9.1}$ & 86.2$_{0.1}$ \\
\rowcolor{xaqamethodbg}
\textbf{X-AQA} & \textbf{60.7}$_{3.6}$ & \textbf{59.2}$_{1.8}$ & \textbf{29.3}$_{3.2}$ & \textbf{89.7}$_{0.1}$ & \textbf{24.4}$_{8.5}$ & \textbf{55.7}$_{5.2}$ & \textbf{12.6}$_{9.9}$ & \textbf{87.1}$_{0.1}$ \\
\midrule
\multicolumn{9}{c}{\textit{Proprietary Models}} \\
Claude Sonnet 5.5 & 32.1$_{4.9}$ & 53.7$_{1.9}$ & 21.0$_{3.4}$ & 85.5$_{0.1}$ & 31.5$_{6.8}$ & 57.2$_{5.2}$ & 15.5$_{10.2}$ & 85.4$_{0.1}$ \\
Gemini 3.8 Flash & 44.6$_{4.4}$ & 52.6$_{1.8}$ & 21.3$_{3.1}$ & 86.7$_{0.1}$ & 30.1$_{7.1}$ & 56.5$_{5.3}$ & 13.2$_{10.4}$ & 86.2$_{0.1}$ \\
GPT-6.1 Sol & 46.5$_{4.9}$ & 56.4$_{1.9}$ & 24.9$_{3.4}$ & 86.2$_{0.1}$ & 19.3$_{7.8}$ & 61.2$_{5.1}$ & 22.4$_{10.2}$ & 85.9$_{0.1}$ \\
\bottomrule
\end{tabularx}
\end{table*}

\subsection{X-AQA-Bench and Evaluation}
\label{sec:eval_protocol}

\paragraph{Benchmark Composition.}
We establish X-AQA-Bench to evaluate assessment performance on seen datasets and generalization to unseen datasets, as illustrated in \cref{fig:training_data} (bottom).
The ID split is constructed from the test sets of datasets represented in X-AQA-40K.
To preserve assessment diversity, we stratify samples within each task according to available action categories or assessment attributes, such as technical points.
Each stratum contains samples of the same action type or questions assessing the same technical point, depending on the task's annotations.
We retain one question per recording and stratum, then select up to 300 samples per task and output format through stratified sampling, accounting for category labels within each stratum when applicable.
The OOD split comprises four held-out datasets: T3Set~\cite{t3set} for textual coaching feedback in competitive sports, Fitness-7~\cite{fitness7} for exercise correctness classification, KIMORE for rehabilitation scoring, and JIGSAWS~\cite{jigsaws} for surgical skill scoring. The four datasets cover all four domains and three assessment formats.
All samples from these datasets are included in the OOD split and excluded from training.
Overall, X-AQA-Bench contains 7,101 evaluation samples across 25 data sources, comprising 5,366 ID and 1,735 OOD samples.

\paragraph{Evaluation Metrics.}
We evaluate three types of AQA tasks using metrics tailored to their output formats.
For continuous scoring, we use Spearman's Rank Correlation Coefficient (SRCC)~\cite{spearman1904proof} to measure the agreement between predicted and reference score rankings.
We form scoring groups containing samples whose scores can meaningfully be compared, such as performances in the same sport or apparatus event within a data source.
We compute SRCC separately within each scoring group and aggregate the correlations using a sample-weighted Fisher $z$ transformation~\cite{fisher1921probable}:
\begin{equation}
    \rho = \tanh\!\left(
        \frac{\sum_{g\in\mathcal{G}} n_g \operatorname{arctanh}(\rho_g)}
             {\sum_{g\in\mathcal{G}} n_g}
    \right)
    \label{eq:srcc_aggregation}
\end{equation}
where $\rho$ is the aggregated SRCC for the evaluated split or domain, and $\rho_g$ and $n_g$ denote the SRCC and number of valid predictions for scoring group $g$, respectively.
The set $\mathcal{G}$ includes all scoring groups with valid predictions.
For categorical assessment, we report macro-F1 over the reference classes and unweighted Cohen's $\kappa$~\cite{cohen1960coefficient}, which accounts for chance agreement.
For textual feedback, we compute BERTScore F1~\cite{zhang2020bertscore} against the reference responses, using RoBERTa-large~\cite{liu2019roberta}.
Categorical and textual metrics are first computed within each data source and then averaged across sources, weighted by the number of valid predictions.
All metrics are computed on successfully parsed predictions.
We also estimate 95\% confidence intervals using 10,000 recording-level bootstrap resamples.

\section{Experiments}
\label{sec:exp}

\begin{table}[t]
\centering
\caption{Comparison with dataset-specific specialists. In Fitness-AQA, KIE and KFE refer to knees-inward and knees-forward errors, respectively. On KIMORE, X-AQA is evaluated in a zero-shot setting, whereas the specialist is trained on the dataset. $\uparrow$ and $\downarrow$ indicate higher and lower are better, respectively.}
\label{tab:specialists}
\small
\begin{tabularx}{\linewidth}{Xcc}
\toprule
\textbf{MTL-AQA} & \textbf{SRCC}$\uparrow$ & \textbf{R-$\ell_2$}$\downarrow$ \\
\midrule
RICA$^2$~\cite{majeedi2024rica2} & 0.9594 & 0.2580 \\
TPT~\cite{bai2022tpt} & \textbf{0.9607} & \textbf{0.2378} \\
DAE-CoRe~\cite{zhang2024dae} & 0.9589 & -- \\
CoRe~\cite{yu2021core} & 0.9512 & 0.2600 \\
TSA-Net~\cite{wang2021tsanet} & 0.9422 & -- \\
MUSDL~\cite{tang2020usdl} & 0.9273 & 0.4510 \\
C3D-AVG-MTL~\cite{mtlaqa} & 0.9044 & -- \\
\midrule
Qwen3.5-9B (base) & 0.6464 & 1.8750 \\
\rowcolor{xaqamethodbg}
\textbf{X-AQA} & 0.8862 & 0.6948 \\
\bottomrule
\end{tabularx}

\par\medskip
\begin{tabularx}{\linewidth}{Xcccc}
\toprule
\multirow{3}{*}{\textbf{Fitness-AQA}} & \multicolumn{4}{c}{\textbf{F1}$\uparrow$} \\
\cmidrule(lr){2-5}
 & \multicolumn{2}{c}{Back Squat} & \multicolumn{2}{c}{Overhead Press} \\
\cmidrule(lr){2-3}\cmidrule(lr){4-5}
 & KIE & KFE & Elbow & Knees \\
\midrule
SimSiam~\cite{chen2021simsiam} & 0.2270 & 0.7868 & 0.4145 & 0.5301 \\
TemporalXform~\cite{jenni2020temporal} & 0.3414 & 0.8319 & 0.4138 & 0.8416 \\
MD~\cite{fitnessaqa} & \textbf{0.4186} & \textbf{0.8338} & \textbf{0.4552} & \textbf{0.8452} \\
\midrule
Qwen3.5-9B (base) & 0.0000 & 0.0000 & 0.0741 & 0.6224 \\
\rowcolor{xaqamethodbg}
\textbf{X-AQA} & 0.2378 & 0.8130 & 0.0714 & 0.6742 \\
\bottomrule
\end{tabularx}

\par\medskip
\begin{tabularx}{\linewidth}{Xccccc}
\toprule
\multirow{2}{*}{\textbf{KIMORE}} & \multicolumn{5}{c}{\textbf{SRCC}$\uparrow$} \\
\cmidrule(lr){2-6}
 & E1 & E2 & E3 & E4 & E5 \\
\midrule
Abedi \etal~\cite{abedi2023crossmodal} & \textbf{0.76} & \textbf{0.61} & \textbf{0.73} & \textbf{0.54} & \textbf{0.67} \\
\midrule
Qwen3.5-9B (base) & 0.00 & 0.00 & 0.00 & 0.00 & 0.00 \\
\rowcolor{xaqamethodbg}
\textbf{X-AQA} & 0.35 & 0.45 & 0.47 & 0.50 & 0.18 \\
\bottomrule
\end{tabularx}

\par\medskip
\begin{tabularx}{\linewidth}{Xcc}
\toprule
\textbf{LASANA} & \textbf{Accuracy}$\uparrow$ & \textbf{Balanced Accuracy}$\uparrow$ \\
\midrule
X3D~\cite{lasana} & \textbf{0.8638} & \textbf{0.8535} \\
\midrule
Qwen3.5-9B (base) & 0.5512 & 0.5669 \\
\rowcolor{xaqamethodbg}
\textbf{X-AQA} & 0.7706 & 0.7576 \\
\bottomrule
\end{tabularx}
\end{table}

\begin{table*}[!t]
\centering
\caption{Effect of multi-domain learning. \textit{Single} uses four independently trained domain-specific models, with each sample evaluated by the corresponding model. \textit{Single (one pass)} trains each model for one epoch, whereas \textit{single (same steps)} repeats its domain-specific training data to match the number of optimization steps used by \textit{joint}. \textit{Joint} trains a single model on data from all four domains. The $\Delta$ row reports \textit{joint} $-$ \textit{single (one pass)}. All reported values are multiplied by 100. Subscripts indicate the half-widths of 95\% bootstrap confidence intervals.}
\label{tab:domain_vs_mixed}
\small
\begin{tabularx}{\textwidth}{Xcccc*{4}{>{\columncolor{xaqamethodbg}}c}}
\toprule
\multirow{3}{*}{\textbf{Method}}
& \multicolumn{4}{c}{\textbf{ID}} & \multicolumn{4}{>{\columncolor{xaqamethodbg}}c}{\textbf{OOD}} \\
\cmidrule(lr){2-5}\cmidrule(lr){6-9}
& \multicolumn{1}{c}{Score} & \multicolumn{2}{c}{Category} & \multicolumn{1}{c}{Text}
& \multicolumn{1}{>{\columncolor{xaqamethodbg}}c}{Score} & \multicolumn{2}{>{\columncolor{xaqamethodbg}}c}{Category} & \multicolumn{1}{>{\columncolor{xaqamethodbg}}c}{Text} \\
\cmidrule(lr){2-2}\cmidrule(lr){3-4}\cmidrule(lr){5-5}\cmidrule(lr){6-6}\cmidrule(lr){7-8}\cmidrule(lr){9-9}
& SRCC$\uparrow$ & macro-F1$\uparrow$ & $\kappa\uparrow$ & BERTScore$\uparrow$
& SRCC$\uparrow$ & macro-F1$\uparrow$ & $\kappa\uparrow$ & BERTScore$\uparrow$ \\
\midrule
Qwen3.5-9B (base) & 21.8$_{5.0}$ & 48.7$_{1.8}$ & 14.1$_{3.3}$ & 87.6$_{0.1}$ & 6.3$_{4.2}$ & 53.1$_{5.3}$ & 6.9$_{10.4}$ & 86.8$_{0.1}$ \\
Single (one pass) & 59.8$_{3.7}$ & 57.7$_{1.8}$ & 26.7$_{3.3}$ & 89.8$_{0.1}$ & 10.7$_{7.7}$ & 49.9$_{5.3}$ & 5.2$_{9.4}$ & 86.8$_{0.1}$ \\
Single (same steps) & \textbf{60.9}$_{3.6}$ & \textbf{60.2}$_{1.8}$ & \textbf{29.9}$_{3.3}$ & \textbf{89.9}$_{0.1}$ & 9.3$_{7.6}$ & 45.8$_{5.1}$ & 4.5$_{7.6}$ & 86.3$_{0.1}$ \\
Joint & 60.7$_{3.6}$ & 59.2$_{1.8}$ & 29.3$_{3.2}$ & 89.7$_{0.1}$ & \textbf{24.4}$_{8.5}$ & \textbf{55.7}$_{5.2}$ & \textbf{12.6}$_{9.9}$ & \textbf{87.1}$_{0.1}$ \\
\midrule
$\Delta$ & +0.9$_{2.9}$ & +1.5$_{1.8}$ & +2.5$_{3.3}$ & -0.1$_{0.1}$ & +13.6$_{8.4}$ & +5.8$_{6.1}$ & +7.5$_{10.5}$ & +0.3$_{0.1}$ \\
\bottomrule
\end{tabularx}
\end{table*}

\begin{table*}[!t]
\centering
\caption{Effect of assessment criteria. For the base model, assessment criteria are included or omitted at inference, while for X-AQA, the corresponding settings are applied during both training and inference. The $\Delta$ row reports \textit{X-AQA (w/ criteria)} $-$ \textit{X-AQA (w/o criteria)}. All reported values are multiplied by 100. Subscripts indicate the half-widths of 95\% bootstrap confidence intervals.}
\label{tab:criteria_ablation}
\small
\begin{tabularx}{\textwidth}{Xcccc*{4}{>{\columncolor{xaqamethodbg}}c}}
\toprule
\multirow{3}{*}{\textbf{Method}}
& \multicolumn{4}{c}{\textbf{ID}} & \multicolumn{4}{>{\columncolor{xaqamethodbg}}c}{\textbf{OOD}} \\
\cmidrule(lr){2-5}\cmidrule(lr){6-9}
& \multicolumn{1}{c}{Score} & \multicolumn{2}{c}{Category} & \multicolumn{1}{c}{Text}
& \multicolumn{1}{>{\columncolor{xaqamethodbg}}c}{Score} & \multicolumn{2}{>{\columncolor{xaqamethodbg}}c}{Category} & \multicolumn{1}{>{\columncolor{xaqamethodbg}}c}{Text} \\
\cmidrule(lr){2-2}\cmidrule(lr){3-4}\cmidrule(lr){5-5}\cmidrule(lr){6-6}\cmidrule(lr){7-8}\cmidrule(lr){9-9}
& SRCC$\uparrow$ & macro-F1$\uparrow$ & $\kappa\uparrow$ & BERTScore$\uparrow$
& SRCC$\uparrow$ & macro-F1$\uparrow$ & $\kappa\uparrow$ & BERTScore$\uparrow$ \\
\midrule
Base (w/o criteria) & 19.2$_{5.5}$ & 45.9$_{1.8}$ & 10.8$_{3.1}$ & 87.7$_{0.1}$ & 2.0$_{7.6}$ & 52.3$_{5.3}$ & 5.2$_{10.4}$ & 86.8$_{0.1}$ \\
Base (w/ criteria) & 21.8$_{5.0}$ & 48.7$_{1.8}$ & 14.1$_{3.3}$ & 87.6$_{0.1}$ & 6.3$_{4.2}$ & 53.1$_{5.3}$ & 6.9$_{10.4}$ & 86.8$_{0.1}$ \\
X-AQA (w/o criteria) & 59.5$_{3.7}$ & 57.1$_{1.8}$ & 26.7$_{3.1}$ & 89.5$_{0.1}$ & 17.7$_{8.4}$ & 47.0$_{5.0}$ & 5.2$_{7.8}$ & \textbf{87.4}$_{0.1}$ \\
X-AQA (w/ criteria) & \textbf{60.7}$_{3.6}$ & \textbf{59.2}$_{1.8}$ & \textbf{29.3}$_{3.2}$ & \textbf{89.7}$_{0.1}$ & \textbf{24.4}$_{8.5}$ & \textbf{55.7}$_{5.2}$ & \textbf{12.6}$_{9.9}$ & 87.1$_{0.1}$ \\
\midrule
$\Delta$ & +1.3$_{2.7}$ & +2.1$_{1.5}$ & +2.6$_{2.7}$ & +0.2$_{0.1}$ & +6.7$_{8.2}$ & +8.7$_{5.6}$ & +7.5$_{10.1}$ & -0.4$_{0.1}$ \\
\bottomrule
\end{tabularx}
\end{table*}

\subsection{Implementation Details}

We fine-tune Qwen3.5-9B using LoRA~\cite{hu2022lora} with a rank of 32, a scaling factor of 64, and a dropout rate of 0.05. LoRA adapters are applied to all linear layers of the language model and visual merger. We use the AdamW~\cite{loshchilov2019adamw} optimizer with a weight decay of 0.1 and gradient clipping at a maximum norm of 1.0. The peak learning rates are $1\times10^{-4}$ and $5\times10^{-5}$ for the language model and visual merger adapters, respectively, following a cosine decay schedule with 3\% warmup and a minimum learning rate of 10\% of the peak value. Training is performed for one epoch on 16 NVIDIA H20 GPUs with a global batch size of 32 and bfloat16 precision. We adaptively sample up to 128 frames per video clip, using higher frame rates for shorter clips and lower rates for longer clips to preserve temporal coverage. The maximum sequence length is 32,768 tokens. Both the training and data random seeds are set to 42. We use the checkpoint from the end of training for evaluation.

\subsection{Main Results}

\paragraph{Comparison with General-Purpose MLLMs.}
We compare X-AQA with both open-source and proprietary MLLMs on X-AQA-Bench.
The open-source models include Molmo2-8B~\cite{clark2026molmo2}, VideoLLaMA3-7B~\cite{zhang2025videollama3}, InternVL3.5-8B~\cite{wang2025internvl35}, Keye-VL-1.5-8B~\cite{yang2025keyevl15}, MiniCPM-V 4.5~\cite{yu2025minicpmv45}, InternVideo3-8B-Instruct~\cite{yan2026internvideo3}, Eagle2.5-8B~\cite{chen2025eagle25}, LLaVA-OneVision-2~\cite{an2026llavaonevision2}, Qwen2.5-VL-7B-Instruct~\cite{bai2025qwen25vl}, Qwen3-VL-8B, and Qwen3.5-9B.
We evaluate both instruct and thinking variants of Qwen3-VL-8B, and both non-thinking and thinking modes of Qwen3.5-9B.
We also include three proprietary models: Claude Sonnet 5.5~\cite{anthropic2026sonnet55}, Gemini 3.8 Flash, and GPT-6.1 Sol.
For categorical assessment, we additionally report two chance-level baselines based on uniform random predictions (random) and the empirical class distribution in the training data (frequency), respectively.
During evaluation, all models use a maximum of 128 frames, except VideoLLaMA3-7B, which is limited to 64 frames due to its context-length constraint.
The results are shown in \cref{tab:benchmark_main}.
X-AQA surpasses all evaluated open-source and proprietary MLLMs on all four ID metrics, demonstrating strong performance across numerical, categorical, and textual assessment.
In particular, it achieves an SRCC of 0.607, compared with 0.465 for GPT-6.1 Sol, the best-performing baseline on this metric.
On the OOD split, X-AQA outperforms all evaluated open-source MLLMs on all four metrics and is comparable with proprietary MLLMs.
For textual feedback, it achieves the highest BERTScore of 0.871, exceeding 0.862 for Gemini 3.8 Flash.
These results demonstrate the effectiveness of unified AQA training for both represented data sources and generalization to unseen datasets.

\paragraph{Comparison with Dataset-Specific Specialists.}
We further compare X-AQA with dataset-specific specialists on four datasets: MTL-AQA, Fitness-AQA, KIMORE, and LASANA.
To facilitate comparison with published specialist results, we evaluate X-AQA on the full test sets rather than the sampled subsets in X-AQA-Bench, using the corresponding evaluation metrics.
None of the test samples are included in X-AQA-40K.
Specifically, we report SRCC and relative $\ell_2$ error for MTL-AQA, error-detection F1 for Fitness-AQA, per-exercise SRCC for KIMORE, and accuracy and balanced accuracy for LASANA.
X-AQA uses the same jointly trained model without dataset-specific fine-tuning.
As shown in \cref{tab:specialists}, dataset-specific specialists generally achieve higher scores on their respective benchmarks, while X-AQA demonstrates competitive assessment performance across diverse tasks with a single model.
Compared with the Qwen3.5-9B base model, X-AQA improves SRCC from 0.6464 to 0.8862 and reduces relative error from 1.8750 to 0.6948 on MTL-AQA.
On LASANA, it increases accuracy from 0.5512 to 0.7706 and balanced accuracy from 0.5669 to 0.7576.
On Fitness-AQA, X-AQA achieves an F1 of 0.8130 for knees-forward error detection, approaching the specialist result of 0.8338.
Beyond these datasets represented during training, X-AQA generalizes to the held-out KIMORE dataset, achieving positive SRCC across all five rehabilitation exercises and reaching 0.50 on E4, compared with 0.54 reported by Abedi \etal~\cite{abedi2023crossmodal}, whose model is trained on KIMORE.
This comparison highlights the distinction between dataset-specific specialization and cross-domain generalization, with X-AQA supporting diverse assessment tasks and actions through a single model.

\subsection{Ablation Study}

\paragraph{Effect of Multi-Domain Learning.}

We investigate the effect of multi-domain learning by comparing multi-domain joint training with four independently trained domain-specific models, each evaluated on samples from its corresponding domain.
We consider two settings for domain-specific training: \textit{single (one pass)}, where each model trains on its domain data for one epoch, and \textit{single (same steps)}, where domain-specific samples are repeatedly drawn to match the number of optimization steps in multi-domain joint training.
As shown in \cref{tab:domain_vs_mixed}, multi-domain learning achieves comparable ID performance while outperforming both single-domain settings on all four OOD metrics.
In particular, multi-domain learning achieves an OOD SRCC of 0.244, compared with 0.107 for \textit{single (one pass)}.
Increasing single-domain training to match multi-domain optimization steps improves ID performance but reduces OOD performance across all four metrics, with SRCC decreasing to 0.093.
These results suggest that AQA tasks across domains share transferable assessment knowledge, enabling multi-domain learning to improve generalization to unseen datasets.

\paragraph{Effect of Assessment Criteria.}

\begin{figure*}
    \centering
    \includegraphics[width=\textwidth]{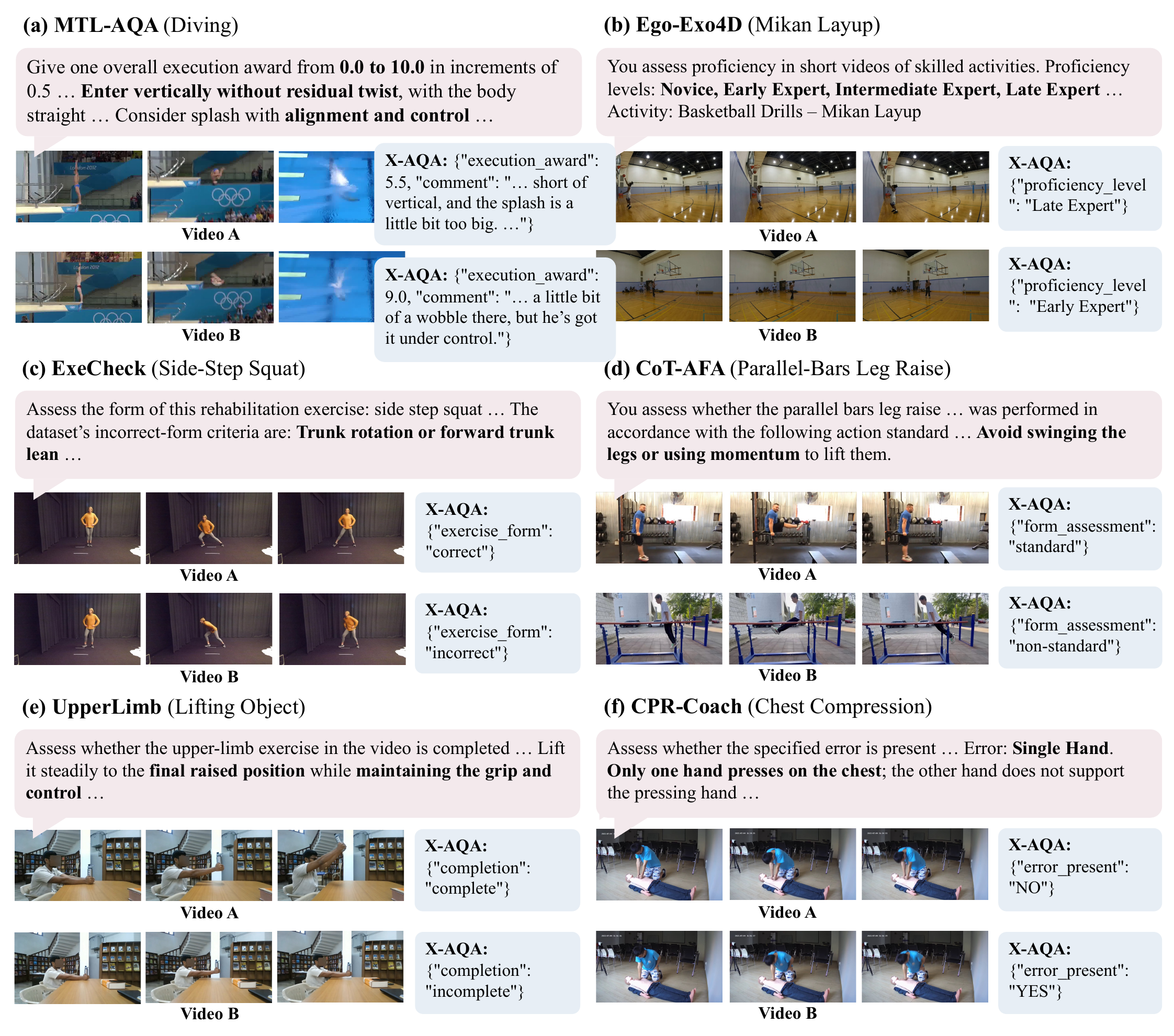}
    \caption{Qualitative examples of X-AQA. In each panel, X-AQA evaluates two different videos using the same prompt and produces distinct assessments results based on the observed action execution and task-specific assessment criteria.}
    \label{fig:qualitative_results}
\end{figure*}

We investigate the effect of incorporating explicit assessment criteria into model prompts for both the base model and X-AQA.
As shown in \cref{tab:criteria_ablation}, providing assessment criteria at inference improves the base model's ID and OOD performance, with OOD SRCC increasing from 0.020 to 0.063.
For X-AQA, incorporating assessment criteria during both training and inference yields further improvements, particularly on unseen datasets.
On the OOD split, SRCC increases from 0.177 to 0.244, macro-F1 from 0.470 to 0.557, and Cohen's $\kappa$ from 0.052 to 0.126.
The ID results also show consistent improvements across all four metrics.
These results indicate that explicit assessment criteria help MLLMs apply task-specific evaluation standards, particularly for numerical scoring and categorical judgments on unseen datasets.

\subsection{Qualitative Results}

\Cref{fig:qualitative_results} presents several examples of qualitative results of X-AQA across diverse action domains and assessment tasks.
Given the same assessment prompt for two different videos, X-AQA produces different judgments based on the observed action execution.
For instance, on MTL-AQA, X-AQA assigns different execution scores to two diving performances and generates feedback describing their execution characteristics in natural language.
The remaining examples demonstrate its ability to assess proficiency levels and exercise correctness, identify execution errors, and determine task completion according to task-specific assessment criteria.
These examples demonstrate X-AQA's ability to perform diverse assessment tasks using assessment criteria within a unified question-answering framework.

\section{Conclusion}
\label{sec:conclusion}

We present X-AQA, the first unified training and evaluation framework for general-purpose AQA with MLLMs.
X-AQA-40K unifies supervision from 22 data sources across four domains, enabling a single model to produce numerical scores, categorical judgments, and textual feedback through question answering.
X-AQA-Bench evaluates assessment performance and generalization to unseen datasets.
X-AQA outperforms all evaluated open-source MLLMs across the benchmark metrics and achieves competitive performance relative to proprietary models.
Ablation studies show that multi-domain learning and explicit assessment criteria improve generalization to unseen datasets.
Our data, benchmark, and model provide a foundation for general-purpose AQA across diverse tasks and domains.

\section*{Acknowledgement}

This work was supported by the Beijing Natural Science Foundation (Grant No. L252171), the Capital's Funds for Health Improvement and Research (Grant No. 2026-2-2191), and Huawei's AI Hundred Schools Program.

\section*{LLM Usage Statement}

We use OpenAI Codex for code implementation and experiment management, including autonomous multi-step workflows under author direction.
We use ChatGPT to polish the language and clarity of author-written manuscript text.
ChatGPT also assists in refining author-prepared related-work and citation materials, figures, and data preparation materials.
The research ideas, hypotheses, scientific claims, and conclusions originate from the authors.
The authors retain full responsibility for the implementation, experimental results, and all content of the paper.

{
    \small
    \bibliographystyle{ieeenat_fullname}
    \bibliography{main}
}

\clearpage
\setcounter{page}{1}
\maketitlesupplementary
\normalsize
\setcounter{section}{0}
\renewcommand{\thesection}{\Alph{section}}
\renewcommand{\theHsection}{supp.\Alph{section}}
\raggedbottom

\section{Dataset Preprocessing}
\label{sec:processing}

In this section, we describe the preprocessing of each source dataset, including video preparation, assessment criteria collection, and QA construction.

\subsection{AQA-7}

AQA-7~\cite{aqa7} contains videos from seven competitive events: individual 10-m platform diving, gymnastics vault, Big Air skiing and snowboarding, synchronized 3-m springboard and 10-m platform diving, and trampoline. Each video is annotated with a total score, with degrees of difficulty additionally provided for individual diving and vault.
We use the released temporally normalized clips and official Split-4 for six sports, excluding trampoline because no split is provided.
For individual diving and vault, where the released scores aggregate assessments from multiple judges, we derive execution score proxies from the total score $S$ and difficulty $D$ as $S/(3D)$ and $S-D$, rounded to the nearest 0.5 and 0.1, respectively.
For Big Air and synchronized diving, we retain the original panel scores.
We select historical scoring rules corresponding to the competition periods represented in AQA-7.
For diving, we follow the January 2010 FINA Diving Officials Manual~\cite{worldaquaticsdivingrules}.
For vault, we use visually observable execution faults from the FIG men's Codes of Points for 2009--2012 and 2013--2016 and women's Codes of Points for 2006--2008, 2009--2012, and 2013--2016~\cite{figrules}.
When the competition year cannot be determined, we retain criteria shared across the relevant editions.
An example QA pair for 10-m platform diving is shown below.

\suppqa{
You are one execution judge at a 10-m platform diving competition.
Give one overall execution award from 0.0 to 10.0 in increments of 0.5.
...
Deduct 0.5-2.0 for insufficient take-off height or confidence,
failure to take off from the end of the platform,
or twisting visibly initiated on the platform.
...
Throughout flight and entry, keep the feet together and toes pointed.
Deduct 0.5-2.0 for flight-position faults.
...
For a head-first entry, extend the arms beyond the head
in line with the body and bring the hands close together.
If one or both arms remain below the head,
the award must not exceed 4.5.
...
Form one overall-impression award using the quality scale
and execution criteria, without assigning equal phase weights.
...
}
{\suppvideo{}\par As one judge, what execution award would you give this dive?}
{\suppjson{
    \suppjsonfield{"execution\_award": 8.5}
}}

\subsection{MTL-AQA}

MTL-AQA~\cite{mtlaqa} contains diving videos from international competitions, annotated with seven judges' scores and broadcast commentary.
We follow the official Split 0 and reconstruct video clips at 25~fps from the released RGB frame sequences.
For numerical scoring, we use the median of the seven raw judge scores, ranging from 0.0 to 10.0 in increments of 0.5.
We clean the commentary by removing scores, rankings, biographical information, and irrelevant broadcast speech, excluding empty or unmatched commentary.
The available scoring and commentary annotations are combined into a single QA sample for each clip.
Similar to AQA-7, we adopt visually observable execution criteria from the FINA rules for 2009--2013 and 2015--2017, selecting the appropriate edition according to the competition~\cite{worldaquaticsdivingrules}.
The assessment prompts specify the dive and its degree of difficulty, while the target score reflects execution quality rather than the competition total.
An example QA pair is shown below.

\suppqa{
You are one execution judge at a 10m platform diving competition.
Give one overall execution award from 0.0 to 10.0 in increments of 0.5.
Give a brief English commentary in one or a few sentences,
speaking naturally as during a sports broadcast.
...
Judge execution independently of difficulty.
The announced dive describes the intended movement;
its difficulty does not earn execution credit.
...
}
{\suppvideo{}\par
Dive: forward 3 1/2 somersaults, pike position
\par Degree of difficulty: 3.0
\par Expected execution: Begin facing the water and rotate forward,
away from the board or platform.
Complete 3 1/2 somersaults with no twists about the body's long axis.
Show a clear pike: bend at the hips while keeping the knees straight,
feet together and toes pointed.
...
\par Assess the execution of this dive.
}
{\suppjson{
    \suppjsonfield{"execution\_award": 8.0,}
    \suppjsonfield{"comment": "It's so graceful, and you see that he's absolutely ripped that entry. There's barely a splash, drags all the bubbles down with him."}
}}

\subsection{Rhythmic Gymnastics}

\begin{figure}
    \centering
    \includegraphics[width=\columnwidth]{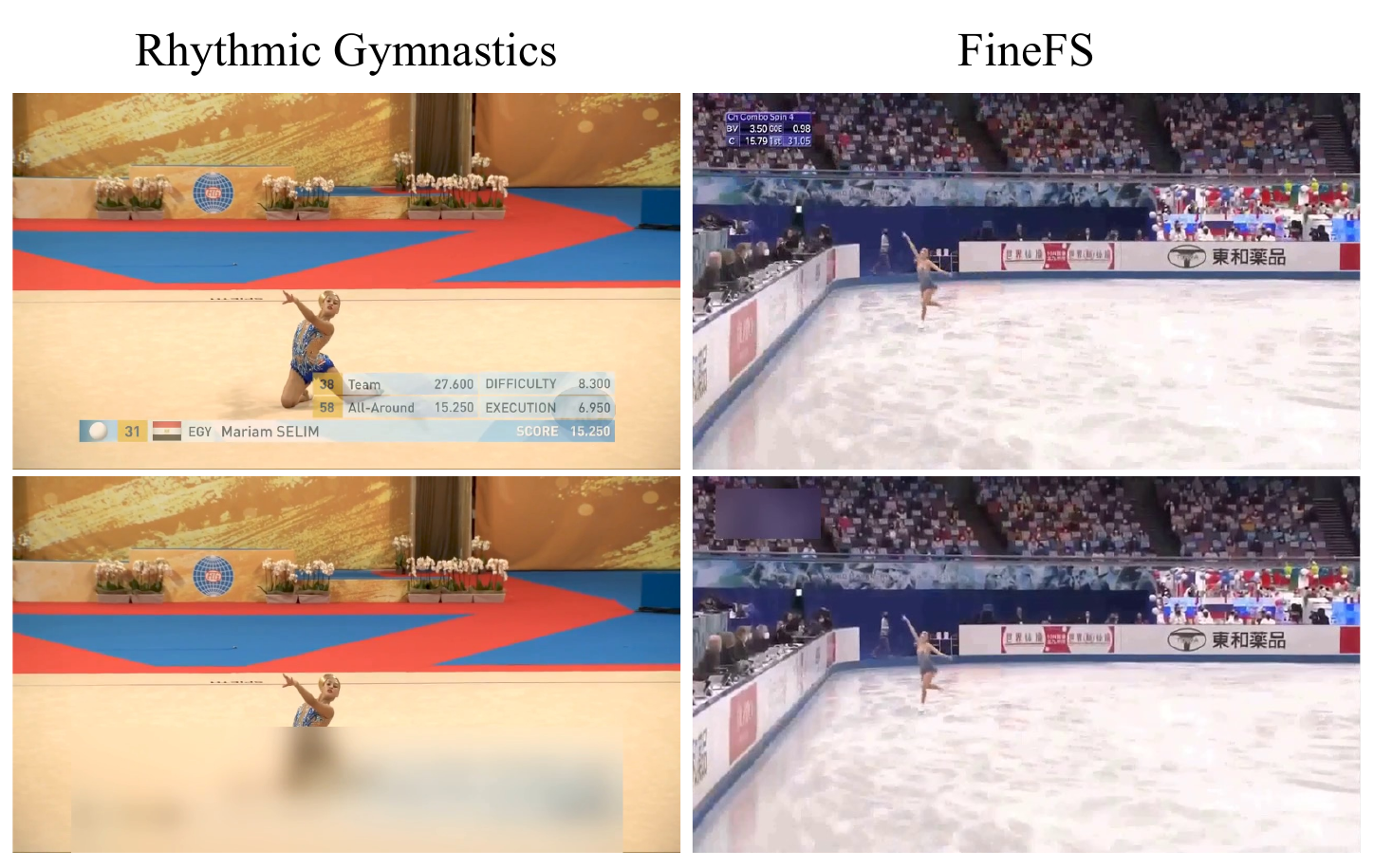}
    \caption{Examples of broadcast scoreboard masking in Rhythmic Gymnastics and FineFS. The top row shows the original video frames, while the bottom row shows the corresponding frames with blurred score panels.}
    \label{fig:score_masking}
\end{figure}

Rhythmic Gymnastics~\cite{rg} contains complete individual routines from FIG competitions, annotated with execution scores and additional penalties for each apparatus.
We retain the released apparatus-wise split and construct two scoring targets for each routine: execution score and additional penalty.
We exclude difficulty scores to focus on execution quality and rule violations rather than routine complexity.
Execution scores are rounded to the nearest 0.1 and penalties to the nearest 0.05.
We incorporate selected apparatus-specific execution and penalty criteria from the FIG Rhythmic Gymnastics Codes of Points for 2013--2016 and 2017--2020, together with the applicable 2018 and 2019 newsletter updates~\cite{figrules}.
The rule version is selected according to each routine's competition identity.
Some videos display broadcast score panels at the bottom of the frame near the end of the routine.
We locate these panels using multi-scale template matching of scoreboard headings and apply Gaussian blur to the detected regions to prevent score leakage.
An example is shown in \cref{fig:score_masking} (left).
The execution scores include both artistic and technical components, while model inputs contain video without audio.
An example for a ball routine is shown below.

\suppqa{
You are a rhythmic gymnastics judge assessing a complete individual routine with ball.
Estimate two separate quantities: execution\_score, which includes artistic and technical execution, and penalty, which contains additional time, boundary and coordinator penalties.
Start execution from 10.00 and subtract artistic and technical deductions; do not award extra execution points for harder elements.
Do not subtract penalty from execution\_score.
...
Routine time is 75-90 seconds, measured from the first movement to final complete stillness; add 0.05 per extra or missing second only when the actual interval is observable.
File length, camera cuts and preparation are not competition time.
Boundary: 0.30 for each body or apparatus occurrence touching the floor outside the boundary.
...
}
{\suppvideo{}\par Apparatus: Ball.
\par Assess this complete individual routine and return its execution score and additional penalty.}
{\suppjson{
\suppjsonfield{"execution\_score": 8.3,}
\suppjsonfield{"penalty": 0.00}
}}

\subsection{FineFS}

FineFS~\cite{finefs} contains figure-skating competition programs with temporally annotated elements and individual judges' grades of execution (GOE).
To limit video length, we use the annotated time intervals to select individual elements from full competition programs.
We take the median of the raw judge grades as the scoring target, ranging from $-5$ to $5$ in increments of 0.5.
This target reflects element-level execution quality rather than converted GOE points or program component scores.
We verify competition identities against official results and use different competitions for training and testing.
Some videos display broadcast score panels in the upper-left corner during the program.
We apply the same scoreboard masking procedure with Rhythmic Gymnastics to prevent score leakage, as shown in \cref{fig:score_masking} (right).
We incorporate element-specific assessment criteria, including positive features, grade limits, and deductions, from the ISU scoring guidelines applicable to each competition season~\cite{isufigureskatingrules}.
We also provide element-specific descriptions of the expected movements.
An example for a Triple Axel is shown below.

\suppqa{
You are a figure skating judge.
Assess the visible execution quality of the specified element and give one GOE grade on a scale from -5 to 5 in half-point increments.
...
Strong elevation and travel distance on every jump in the element.
Controlled take-off followed by a secure landing.
...
Grades +4 and +5 require strong elevation and distance, controlled take-off and landing, and easy, fluent execution throughout; otherwise cap the starting grade at +3.
...
Fall: subtract 5.
Two-foot landing or a step-out used to recover the landing: subtract 3 to 4.
...
Subtract the total deduction from the capped starting grade, then clamp to [-5, +5].
...
}
{\suppvideo{}\par
Competition: 2019 World Championships; Men's singles; Short program; season 2018/19
\par Element: 3A
\par Action: Triple Axel
\par Expected execution: Take off from a forward outside edge without toe-pick assistance and land backward on the outside edge of the other foot.
This Axel has 3 1/2 rotations from forward take-off to backward landing.
Control the body position through rotation and the landing, then continue with a stable, flowing exit.
Score the execution quality of this element.
}
{\suppjson{
    \suppjsonfield{"goe\_grade": 2.0}
}}

\subsection{MultiSenseBadminton}

MultiSenseBadminton~\cite{multisensebadminton} contains recordings of 25 participants performing two badminton strokes, forehand clear and backhand drive, with skill-level annotations and coaching feedback from three coaches.
To accommodate long recordings, we divide the front-view videos into 32-second windows, each inheriting the original participant--stroke skill level and merged coaching feedback.
We adopt the first leave-three-out participant split from the original paper, using 22 participants for training and 3 for testing.
We formulate assessment criteria covering movement technique and shot quality based on the aspects represented in the source annotations.
An example for backhand drive is shown below.

\suppqa{
You are a badminton coach assessing a player's overall skill in the stroke specified by the user.
Base the assessment on the performance visible in the video.
Assessment criteria: Evaluate movement technique: grip, posture, swing mechanics, coordination of the torso, arm and wrist, footwork, and contact point.
Evaluate shot quality: accuracy, distance, direction, speed, and consistency across the visible attempts.
Use only criteria that the camera view allows you to assess; an obscured detail is not evidence of an execution error.
Classify the player's skill level as Beginner, Intermediate, or Expert.
Briefly explain the observed strengths and weaknesses and give relevant execution advice.
...
}
{\suppvideo{}\par Assess the player's overall backhand drive skill level and provide coaching feedback.}
{\suppjson{
    \suppjsonfield{"skill\_level": "Beginner",}
    \suppjsonfield{"feedback": "The basic posture and grip are poor, and there is no backswing."}
}}

\subsection{Ego-Exo4D}

Ego-Exo4D~\cite{egoexo4d} contains egocentric and exocentric recordings of skilled activities, annotated with demonstrator proficiency levels for each take.
We select the designated best exocentric view and divide each recording into 32-second windows, with the final window aligned to the recording end.
All windows from the same take inherit its original proficiency label.
We follow the official train and validation split, using validation takes for testing because the official test labels are not publicly available.
We assign basketball, soccer, and rock climbing to competitive sports, and cooking, music, and dance to general skills.
Both domains share the same QA format, using the four original proficiency levels without additional assessment criteria.
An example for soccer dribbling is shown below.

\suppqa{
You assess proficiency in short videos of skilled activities.
Proficiency levels: Novice, Early Expert, Intermediate Expert, Late Expert
...
}
{\suppvideo{}\par Activity: Soccer Drills - Dribbling}
{\suppjson{
    \suppjsonfield{"proficiency\_level": "Intermediate Expert"}
}}

\subsection{T3Set}

T3Set~\cite{t3set} contains table-tennis practice rounds with coaching feedback from two coaches across different stroke techniques.
We use the complete practice-round videos, selecting the remote view when available and the right view otherwise.
We consolidate each coach's feedback into an English paragraph and construct a separate QA sample for each coach, excluding empty feedback and annotations requiring further review.
The assessment prompt specifies the stroke technique and requests actionable coaching feedback without additional scoring criteria.
We combine the existing round-level train, validation, and test partitions for OOD evaluation.
An example for forehand topspin is shown below.

\suppqa{
You are a table tennis coach.
Assess the player's execution from the video and give specific, actionable feedback.
Address the player directly in the second person.
Respond in a few concise English sentences within a single paragraph.
}
{\suppvideo{}\par Technique: forehand topspin (topspin).
Assess the player's execution across this practice round and provide coaching feedback.}
{Your center of gravity is too high, and the wrist and upper arm direct too much force upward. ...}

\subsection{CricketVision}

CricketVision~\cite{ujaqa} contains cricket batting videos annotated with execution scores for five body parts: head, shoulder, hands, hips, and feet.
We combine the temporally adjacent Buildup, Execution, and FollowThrough phases into a complete batting stroke, retaining samples with valid execution scores for all five body parts.
The model observes the complete stroke and predicts five integer scores from 1 to 10 for the Execution phase, decoded from the original annotation labels.
We formulate assessment criteria based on the body-part alignment and deviation guidelines in the original paper and its supplementary scoring table.
We split the data by source recording, keeping all strokes from the same recording in the same split.
An example is shown below.

\suppqa{
You assess cricket batting technique.
The video shows one batting action.
Score the Execution phase, around the instant immediately before bat-ball contact; use the preparation and follow-through only as context.
Assess these five body parts independently:
1. Head: judge how well the head aligns with the front foot and the reference alignment through the head. Greater angular departure lowers quality.
2. Shoulder: judge the side-on stance and alignment of the two shoulders. Greater angular departure from the appropriate alignment lowers quality.
3. Hands: judge hand placement relative to the vertical line beneath the head and eyes. Greater displacement from the appropriate position lowers quality.
...
Apply the alignment appropriate to the stroke: on-drive and square-drive/cut can require different shoulder alignment; square-drive/cut can require different hand placement; cut and hook allow different hip orientation.
...
}
{\suppvideo{}\par Assess the Execution phase of this cricket batting action.}
{\suppjson{
    \suppjsonfield{"head": 6,}
    \suppjsonfield{"shoulder": 6,}
    \suppjsonfield{"hands": 5,}
    \suppjsonfield{"hips": 6,}
    \suppjsonfield{"feet": 7}
}}

\subsection{ExpertAF}

ExpertAF~\cite{expertaf} provides four-second clips of skilled activities from Ego-Exo4D, paired with concise, actionable feedback generated by summarizing expert commentary.
We select the annotated four-second intervals from the designated best exocentric view.
For training, we remove duplicate entries arising from multiple body-part references and retrieved expert demonstrations, while retaining all released validation samples for testing.
We adapt the original task to single-video feedback generation, using the exocentric video without pose inputs or retrieved demonstrations.
The prompt requests a single actionable coaching suggestion without additional assessment criteria.
An example for rock climbing is shown below.

\suppqa{
You analyze videos of skilled activities and give one concise, actionable coaching suggestion in a single English sentence.
Write the feedback in the third person.
}
{\suppvideo{}\par Activity: Rock Climbing
\par Give feedback on the execution.}
{The climber needs to maintain pressure on the right foot and coordinate a layback move with hand placement, hip rise, and shoulder throw to successfully execute the move.}

\subsection{CoT-AFA}

CoT-AFA~\cite{cotafa} provides action videos of martial arts and fitness exercises, which we assign to competitive sports and fitness exercises, respectively.
The dataset includes human-annotated form labels, explanatory captions, and category-specific reference forms consisting of five action steps and a general instruction.
We incorporate these reference forms into the assessment prompts and construct two QA tasks: standard/non-standard classification based on human labels and explanatory feedback generation using available captions.
Both tasks use complete video clips and follow the released train/test split.
Examples of explanatory feedback for cloud hands and standard/non-standard classification for a pull-up are shown below.

\suppqa{
You assess whether the pull up (a pull-up bar exercise) in the given video was performed in accordance with the following action standard.
...
Step 3: The player should begin from a dead hang, with arms fully extended and shoulders relaxed (but engaged).
Avoid swinging or kipping; the movement should be controlled.
...
}
{\suppvideo{}}
{\suppjson{
    \suppjsonfield{"form\_assessment": "standard"}
}}

\suppqa{
You assess whether the cloud hands (a Tai chi quan technique) in the given video was performed in accordance with the following action standard.
Step 1: The player typically begins from a neutral stance or a transitional posture within the Tai Chi form.
The body should be relaxed, centered, and the weight evenly distributed, with knees slightly bent.
...
Open with one sentence naming the action and stating whether the execution is good or bad, then explain the judgement in a few sentences: what the performer does, which reference steps the execution does or does not follow, and how to correct any error.
Respond in a single paragraph.
}
{\suppvideo{}}
{The cloud hands action in this video is bad. The character starts with an incorrect stance, as they are not centered and balanced. Their weight distribution is uneven, and their knees are not bent enough. Additionally, the hand movements are not coordinated and circular; instead, they appear jerky and disjointed. To improve, the character should begin by centering themselves and distributing their weight evenly, bending their knees slightly. They should also practice smooth, continuous hand movements that flow together, focusing on the waist's coordination with the weight shift.}

\subsection{Fitness-AQA}

Fitness-AQA~\cite{fitnessaqa} contains videos of squats and overhead presses annotated with execution errors at different technical points.
We construct binary QA tasks by expressing each error category as a desired form condition, with ``yes'' indicating the absence of the corresponding error.
For overhead presses and squats, we use complete videos and determine error presence from the annotated temporal intervals.
For shallow squats, we extract the annotated frames as image inputs to preserve frame-level supervision.
We combine the released training and validation splits for training and retain the original test split.
An example for the overhead press is shown below.

\suppqa{
You evaluate whether an exercise execution satisfies a given technical point.
Return only this JSON object and nothing else: \{"technical\_point\_satisfied": "yes" or "no"\}.
}
{\suppvideo{}\par Exercise: Overhead Press
\par Technical point: Keep the elbows in the correct position under the bar, without flaring them out.}
{\suppjson{
    \suppjsonfield{"technical\_point\_satisfied": "yes"}
}}

\subsection{FLAG3D}

FLAG3D~\cite{flag3d} provides 3D-annotated fitness exercises, while FLAG3D++~\cite{flag3dpp} introduces language-based action standards and repetition-level quality annotations.
We use RGB avatar renderings and segment each exercise sequence into individual repetitions according to the annotated temporal boundaries.
Each repetition is paired with the corresponding technical points from the released annotations to construct binary QA samples.
We hold out participants P009 and P010 for testing.
The validation recordings are excluded.
An example for Svend Press is shown below.

\suppqa{
You evaluate whether an exercise execution satisfies a given technical point.
Return only this JSON object and nothing else: \{"technical\_point\_satisfied": "yes" or "no"\}.
}
{\suppvideo{}\par Exercise: Svend Press
\par Technical point: stand shoulder-width apart}
{\suppjson{
    \suppjsonfield{"technical\_point\_satisfied": "yes"}
}}

\subsection{EgoExo-Fitness}

EgoExo-Fitness~\cite{egoexofitness} provides full-body exercise recordings annotated with technical-point judgments and overall coaching feedback.
We reconstruct action clips at 30~fps from the released frame sequences and randomly select one exocentric view per action instance.
We split the data by participant, keeping all recordings of each participant in the same split.
The assessment criteria are taken directly from the technical-point descriptions in the released interpretable-action annotations.
We construct two QA tasks: binary assessment of individual technical points and generation of overall coaching feedback.
For the latter, the prompt provides all technical points without their reference judgments, allowing the model to assess execution and generate feedback from the video.
Examples of both tasks for kneeling push-ups are shown below.

\suppqa{
You evaluate whether an exercise execution satisfies a given technical point.
Return only this JSON object and nothing else: \{"technical\_point\_satisfied": "yes" or "no"\}.
}
{\suppvideo{}\par Exercise: Kneeling pushing-ups
\par Technical point: Cross your feet.}
{\suppjson{
    \suppjsonfield{"technical\_point\_satisfied": "yes"}
}}

\suppqa{
You give concise, concrete feedback on exercise technique.
Assess the execution visible in the video against the provided technical points.
Respond in one or two sentences in the second person.
}
{\suppvideo{}\par Exercise: Kneeling pushing-ups
\par Technical points:
\par Cross your feet.
\par Keep your back straight.
...
Comment on the execution.}
{The movements were generally in accordance with the instructional text, performed at a fast pace, and were executed smoothly. However, the back still needs to be straighter, so next time try to keep the back even straighter.}

\subsection{QEVD}

QEVD~\cite{qevd} provides two fitness-coaching datasets: FIT-300K for video question answering and FIT-COACH for timestamped coaching feedback.
FIT-300K includes high-level questions about exercise identity and overall execution, and fine-grained questions about specific movements and aspects of exercise form.
We exclude questions solely about action recognition and retain one question per clip.
For FIT-COACH, we construct a QA sample at each feedback timestamp, using up to eight preceding seconds of video and the timestamped history of earlier coaching feedback within the same exercise segment.
We use FIT-300K exclusively for training, while FIT-COACH provides training data and its benchmark subset serves as the test set.
Both tasks use the released QA and coaching annotations without additional assessment criteria.
Examples of high-level QA, fine-grained QA, and coaching feedback are shown below.

\suppqa{
You analyze workout videos and briefly describe the exercise and its execution.
Respond in one to three concise sentences.
}
{\suppvideo{}\par What exercise is the user doing, and how are they performing it?}
{The user is doing a downward dog pose with their body facing the front. However, they need to work on pressing their heels down towards the floor.}

\suppqa{
You answer specific questions about visible movements in workout videos.
Respond in one to three concise sentences.
Begin with Yes or No.
}
{\suppvideo{}\par Are the user's knees together?}
{No, the user's knees are not together.}

\suppqa{
You are a workout coach.
Give one brief, timely, and useful comment based on the exercise shown so far.
}
{\suppvideo{}\par Previous feedback:
\par \textless{}0.0 seconds\textgreater{} Moving on to jumping jacks!
\par \textless{}4.8 seconds\textgreater{} Good job!
\par \textless{}9.1 seconds\textgreater{} 5 reps!
\par \textless{}12.3 seconds\textgreater{} You are not moving your hands correctly.
\par \textless{}15.8 seconds\textgreater{} You gotta move your arms too! I wanna see you moving those arms.
\par \textless{}19.7 seconds\textgreater{} Yes! That is awesome.
\par What would you say now?}
{Now you stopped jumping! Consistency is key to success.}

\subsection{Fitness-7}

Fitness-7~\cite{fitness7} contains videos of seven resistance exercises performed by 26 participants, recorded from three camera views and annotated with good/bad form labels.
We use the complete front-view videos and their original labels, retaining one video per execution.
We hold out five participants for testing.
Each QA sample asks the model to assess overall exercise form using the exercise name and video, without additional assessment criteria.
An example for the standing dumbbell side bend is shown below.

\suppqa{
Assess the exercise form in the video.
\par Exercise: Standing dumbbell side bend.
\par Determine whether the demonstrated exercise form is good or bad.
\par Return only this JSON object and nothing else: \{"exercise\_form": "good" or "bad"\}.
}
{\suppvideo{}}
{\suppjson{
    \suppjsonfield{"exercise\_form": "bad"}
}}

\subsection{KIMORE}

KIMORE~\cite{capecci2019kimore,capecci2018hsmm,capecci2018instrumental} contains recordings of five low-back-pain rehabilitation exercises performed by 78 participants, with clinical assessments based on the ten-item Exercise Accuracy Assessment Questionnaire (EAAQ).
We provide the exercise-specific movement instructions and the ten EAAQ items as assessment criteria, asking the model to rate each item on a five-point scale from Never to Always.
The released clinical annotations provide the clinical Total Score (cTS) rather than individual item ratings.
We therefore use cTS as the reference score and sum the ten predicted item ratings to obtain the model's overall score.
All recordings with available RGB videos and cTS annotations are reserved for OOD evaluation.
An example for the squat exercise is shown below.

\suppqa{
You are a physiotherapy examiner. Evaluate the rehabilitation exercise in the video.
The exercise is Squat.
The subject holds the arms at ninety degrees to the trunk with the elbows extended, flexes the knees up to 60--70 degrees and returns to the starting position, keeping the body aligned in the sagittal plane.
Rate the performance on the ten items of the Exercise Accuracy Assessment Questionnaire (EAAQ), using the scale 1 = Never, 2 = Rarely, 3 = Sometimes, 4 = Often, 5 = Always.
Consider the whole exercise, all repetitions, and judge only what is visible in the video.
- primary\_objective\_achieved: Is the primary objective of the exercise achieved (i.e., extension of the upper limbs, trunk rotation with upper limbs elevated to 90 degrees, squatting, etc.)?
- repetitions\_constant: Is the exercise repeated in a constant manner?
- movement\_amplitude\_complete: Is the amplitude of the movement complete?
...
Return only this JSON object and nothing else, with one score for every item:
...
The clinical Total Score is the sum of the ten scores.
}
{\suppvideo{}}
{\suppjson{
    \suppjsonfield{"cTS": 48.333}
}}

\subsection{KERAAL}

KERAAL~\cite{keraal} contains recordings of three low-back-pain rehabilitation exercises, annotated with correct execution and exercise-specific movement errors.
We resize and letterbox the RGB videos to a consistent resolution and divide longer recordings into windows of up to 32 seconds, each inheriting the recording-level label.
We formulate a four-class classification task using the correct-execution label and three exercise-specific error categories, with descriptions adapted from the dataset paper.
The four answer options are shuffled deterministically for each recording.
We use Group~3 for training and exclude Groups~1A and 2A, as their videos contain robot-rendered skeleton overlays.
An example for the hiding face exercise is shown below.

\suppqa{
You classify how a low-back-pain rehabilitation exercise was performed.
Return only this JSON object and nothing else: \{"answer": "\textless{}selected option letter\textgreater{}"\}.
The value must be one of the option letters shown for the current sample.
}
{\suppvideo{}\par Exercise: hiding face - a breathing exercise with the upper limbs flexed 90 degrees at the shoulder and at the elbow.
\par Possible verdicts:
\par A. the elbows are not bent enough
\par B. the arms are not outspread enough
\par C. correct execution
\par D. the arms are not raised enough}
{\suppjson{
    \suppjsonfield{"answer": "C"}
}}

\subsection{REHAB24-6}

REHAB24-6~\cite{rehab24paper,rehab24} contains multi-view recordings of six rehabilitation exercises, with correctness labels for individual repetitions and a general movement description for each exercise type.
We use Camera~17 RGB videos, resized while preserving the original frame indices, and select individual repetitions according to their annotated temporal boundaries.
Each repetition is formulated as a binary correctness assessment using the released labels.
Based on the dataset's exercise descriptions, we formulate exercise-specific assessment criteria based on these movement descriptions, specifying observable requirements for correct execution.
These criteria are included in the prompts, while supervision is provided by the original repetition-level labels.
We hold out participants 2, 3, and 6 for testing.
An example for arm abduction is shown below.

\suppqa{
You judge how well one repetition of a rehabilitation exercise was performed, as a physiotherapist would.
Exercise: Arm abduction (sideway raising of the straightened right arm)
Performed with: right arm
A correct repetition meets every one of these criteria:
1. The arm stays straight at the elbow through the whole repetition.
2. The arm travels sideways in the frontal plane, not forward across the body.
3. The arm is raised at least to shoulder level.
4. The trunk stays upright, without leaning to the opposite side or shrugging the shoulder.
5. The arm is lowered back to the side under control.
Return only this JSON object and nothing else: \{"rehabilitation\_form": "correct" or "incorrect"\}.
}
{\suppvideo{}}
{\suppjson{
    \suppjsonfield{"rehabilitation\_form": "correct"}
}}

\subsection{ExeCheck}

ExeCheck~\cite{execheck} contains 140 recordings of ten rehabilitation exercises performed by seven participants, each recording comprising five repetitions with a correct/incorrect execution label.
We extract the RGB streams and select individual repetitions according to the released temporal boundaries.
Each repetition inherits the correctness label of its source recording.
We incorporate exercise-specific assessment criteria adapted from the incorrect-form descriptions in its paper.
Following the original evaluation protocol, we use the first six participants for training and hold out the seventh for testing.
An example for arm circle is shown below.

\suppqa{
Assess the form of this rehabilitation exercise: arm circle.
The dataset's incorrect-form criteria are: Raised shoulders or bent arms.
Judge the visible repetition against these criteria.
Incorrect means at least one listed form error; correct means the intended form is maintained.
Return only this JSON object and nothing else: \{"exercise\_form": "correct" or "incorrect"\}.
}
{\suppvideo{}}
{\suppjson{
    \suppjsonfield{"exercise\_form": "correct"}
}}

\subsection{LLM-FMS}

LLM-FMS~\cite{llmfms} provides RGB keyframes of 15 Functional Movement Screen (FMS) movement variants, annotated with item-level judgments and overall quality scores.
We construct one QA sample per keyframe, requiring the model to assess each item and predict the overall score.
The assessment criteria, including movement-specific definitions, thresholds, and scoring rules, are taken from the supplementary S3 Text of the source paper.
We retain the released item-level judgments and align the overall scores with the provided scoring rules where they determine the score, retaining the original annotations otherwise.
We hold out every fifth participant for testing, keeping all movements and recording episodes of each participant in the same split.
An example for the deep squat is shown below.

\suppqa{
You assess the quality of the Functional Movement Screen (FMS) movement "Deep Squat (Floor)" from a single keyframe.
Judge every item below, then give the overall score.

Trunk-calf angle -- The angle between the trunk (the line from the midpoint of the hips to the midpoint of the shoulders) and the tibia (the line from the knee to the ankle).
Parallel: standard, 0-5 degrees
Not parallel: non-standard, over 5 degrees
...
Overall score:
3 = every item above is standard
2 = the squat is completed but at least one item is non-standard
1 = the squat cannot be completed on the floor
...
}
{\suppimage{}}
{\suppjson{
\suppjsonfield{"Trunk-calf angle": "Parallel",}
\suppjsonfield{"Hip height": "Higher",}
\suppjsonfield{"Wrist position": "The Left",}
\suppjsonfield{"Score": 2}
}}

\subsection{UpperLimb}

UpperLimb~\cite{upperlimb} contains RGB recordings of four upper-limb rehabilitation exercises, annotated with complete/incomplete execution labels.
We use the videos and the completion labels to construct QA samples.
Based on the exercise descriptions and completion definitions in the source paper, we formulate assessment criteria specifying the expected movement procedures and endpoints.
The prompt provides the procedures for all four exercises, requiring the model to identify the demonstrated movement and assess its completion.
We use the official train/test split.
An example of QA pair is shown below.

\suppqa{
Assess whether the upper-limb exercise in the video is completed.
Exercise candidates and expected procedures:
...
Extending the Elbow: Begin with the elbow supported at the table and the forearm upright.
Extend the elbow smoothly until the forearm lies along and contacts the table.
Failure to achieve that endpoint or an early stop indicates incomplete execution.
...
Apply the procedure for the exercise shown in the video.
Complete means the required movement reaches its endpoint successfully; incomplete means it visibly fails or stops short.
Assess the recorded cycle without requiring repeated cycles or exact joint angles.
Do not infer unseen failure from occlusion alone.
Return only one JSON object and nothing else, without Markdown or code fences: {"completion": "complete" or "incomplete"}.
}
{\suppvideo{}}
{\suppjson{
\suppjsonfield{"completion": "incomplete"}
}}

\subsection{JIGSAWS}

JIGSAWS~\cite{jigsaws} contains robotic surgical training trials covering suturing, knot tying, and needle passing, annotated with six modified-OSATS skill ratings.
We divide each trial into windows of up to 32 seconds, with each window inheriting the trial-level ratings.
The assessment criteria follow the six dimensions and their scoring anchors at levels 1, 3, and 5 from the source paper, with intermediate scores used for performance between adjacent anchors.
We also provide task-specific procedure descriptions based on the dataset documentation.
An example for knot tying is shown below.

\suppqa{
You are a surgical skills examiner.
Evaluate the robot-assisted surgical training trial provided by the user.
The video may be an excerpt of a longer trial; score the trial it comes from.
Score the performance using the six named dimensions of the modified OSATS global rating score.
...
Respect for tissue:

Lowest anchor (score value 1): Frequently used unnecessary force on tissue

Middle anchor (score value 3): Careful tissue handling but occasionally caused inadvertent damage

Highest anchor (score value 5): Consistent appropriate tissue handling
...
}
{Assess the trial in the following video.
\par Task: Knot Tying
\par Expected procedure: Using the robotic instruments, the subject ties two sutures, one after the other, around a flexible tube fixed at both ends. Each completed knot should be taut and secure around the tube.
\par \suppvideo{}}
{\suppjson{
\suppjsonfield{"Respect for tissue": 2,}
\suppjsonfield{"Suture/needle handling": 2,}
\suppjsonfield{"Time and motion": 2,}
\suppjsonfield{"Flow of operation": 2,}
\suppjsonfield{"Overall performance": 2,}
\suppjsonfield{"Quality of final product": 3}
}}

\subsection{CaptainCook4D}

CaptainCook4D~\cite{captaincook4d} contains cooking recordings annotated with step-level errors across eight categories.
We select individual cooking steps according to their annotated temporal boundaries and provide the recipe and intended step as context.
Each QA sample requires the model to identify execution errors and generate a brief description for each applicable category, returning an empty JSON object when no error is present.
The assessment criteria are adapted from the dataset's error taxonomy, with task dependencies used to distinguish valid alternative step orders from ordering errors.
When multiple errors belong to the same category, their descriptions are consolidated into a single response.
We retain the official subject-level train/test split and exclude validation data.
An example for the pinwheels recipe is shown below.

\suppqa{
Assess the execution of the specified cooking step in the video.
The input video shows a segment of a single cooking step.
Assessment criteria:
Use the intended step as the reference for ingredients, quantities, utensils, preparation, technique, timing and temperature.
Compare the visible actions and outcomes with these requirements to identify deviations.
Alternative step orders that satisfy the required dependencies are valid.
...
Error categories:
...
Order Error: Execution of steps in a sequence that violates the required dependencies.
...
Return {} when there are no errors.
}
{\suppvideo{}\par Recipe: Pinwheels
\par Intended step: scoop-Use a butter knife to scoop nut butter from the jar
\par Assess this step and return the error descriptions by category.}
{\suppjson{
\suppjsonfield{"Order Error": "Step performed before placing tortilla on cutting board"}
}}

\subsection{AIxSuture}

AIxSuture~\cite{hoffmann2024aixsuture} contains recordings of open suturing on a bench model, assessed by three raters across eight OSATS skill dimensions.
We divide each recording into windows of up to 32 seconds, with each window inheriting the recording-level ratings.
The assessment criteria follow the eight OSATS dimensions and their scoring anchors at levels 1, 3, and 5.
For each dimension, we use the median of the three raters' scores as the reference, and sum the eight dimension scores to obtain the global rating score (GRS).
We retain the released subject-level train/test split, excluding the validation data.
An example of suturing assessment is shown below.

\suppqa{
You are a surgical skills examiner.
Evaluate the open suturing performance in the video.
The subject performs single interrupted sutures on an ethylene-vinyl acetate bench model with a needle holder, forceps and scissors; judge only the performance visible in the video.
Score the performance on the eight dimensions of the Objective Structured Assessment of Technical Skills (OSATS) global rating score.
Apply each dimension to the suturing task and use a five-point Likert scale.
...
Respect for tissue:

Anchor 1: Often used unnecessary force on tissue or caused damage by inappropriate use of instruments

Anchor 3: Careful handling of tissue but occasionally caused inadvertent damage

Anchor 5: Consistently handled tissues appropriately with minimal damage
...
}
{\suppvideo{}\par Assess the suturing performance in the following video.}
{\suppjson{
\suppjsonfield{"Respect for tissue": 1,}
\suppjsonfield{"Time and motion": 2,}
\suppjsonfield{"Instrument handling": 1,}
\suppjsonfield{"Suture technique": 2,}
\suppjsonfield{"Flow of operation and forward planning": 2,}
\suppjsonfield{"Knowledge of specific procedure": 2,}
\suppjsonfield{"Quality of final product": 1,}
\suppjsonfield{"Overall performance": 2}
}}

\subsection{CPR-Coach}

CPR-Coach~\cite{cprcoach} contains chest-compression training recordings from four synchronized camera views, annotated with execution errors.
We reconstruct RGB videos at 25~fps from the released frame sequences and use camera ch0 for all recordings.
We construct binary QA samples for each annotated error and sample an equal number of absent error categories, using one absent-error question for recordings with correct execution.
To provide balanced positive and negative supervision for error detection, we construct binary QA samples for each annotated error and an equal number of randomly sampled absent error categories. For recordings with correct execution, we sample one absent-error category.
We retain the official train/test split, including recordings with multiple simultaneous errors in the test set.
An example for hand placement is shown below.

\suppqa{
Assess whether the specified error is present in the chest-compression action.
Return only this JSON object and nothing else: \{"error\_present": "YES" or "NO"\}.
}
{\suppvideo{}\par Error: Overlap Hands
\par Description: The hands are laid flat across one another with the fingers extended and not interlocked during compression. Look at the finger arrangement; having one hand above the other alone does not identify this error.}
{\suppjson{
\suppjsonfield{"error\_present": "YES"}
}}

\subsection{LASANA}

LASANA~\cite{lasana} contains laparoscopic training recordings with four skill-dimension ratings and task-specific error annotations.
We use left-camera videos and construct two QA tasks: numerical assessment of surgical skill and binary detection of individual execution errors.
The scoring task evaluates depth perception, efficiency, bimanual dexterity, and tissue handling, using the median of the available raw ratings from at least two raters for each dimension.
The error-detection task asks whether a specified task-specific error occurs in the video.
We adapt the scoring anchors and error definitions from the source paper.
We retain the official train/test split and exclude validation data.
Examples of skill scoring and error detection are shown below.

\suppqa{
Assess the four skill dimensions over the complete laparoscopic training video.
Assess each dimension independently from 1.0 to 5.0 in increments of 0.5 using the task-specific anchors below.
Use intermediate values for performance between adjacent anchors.
Scoring criteria:

Depth perception: accuracy of instrument targeting, grasping and placement.
1: Missed grasps; 3: Slow accuracy; 5: Fast precise transfers.
...
}
{\suppvideo{}\par Task: Peg transfer
\par Assess the four skill dimensions.}
{\suppjson{
\suppjsonfield{"Depth perception": 4.0,}
\suppjsonfield{"Efficiency": 4.0,}
\suppjsonfield{"Bimanual dexterity": 4.0,}
\suppjsonfield{"Tissue handling": 4.0}
}}

\suppqa{
Determine whether the specified error occurs in the laparoscopic training video.
Return only this JSON object and nothing else: \{"error\_present": "Yes" or "No"\}.
}
{\suppvideo{}\par Task: Balloon resection
\par Error: balloon\_damaged
\par Definition: The inner balloon is damaged and loses less than half its water.
\par Does this error occur?}
{\suppjson{
\suppjsonfield{"error\_present": "No"}
}}

\section{Dataset Statistics}
\label{sec:supp_statistics}

\subsection{Task-Level Composition}

\Cref{tab:dataset} presents the task-level composition of X-AQA-40K and X-AQA-Bench.
X-AQA-40K contains 40,163 training samples drawn from 22 source datasets across four domains: competitive sports, fitness exercises, rehabilitation, and general skills.
The source counts indicate the available QA samples after preprocessing, while the X-AQA-40K column reports the samples selected for joint training.
These datasets contribute 29 task-specific training subsets, each capped at 2,000 samples.
For Ego-Exo4D and CoT-AFA, which each span two domains, we apply this cap separately to each task within each domain.
X-AQA-Bench comprises 7,101 evaluation samples, including 5,366 ID and 1,735 OOD samples.
For ID evaluation, we select up to 300 samples per task and output format from the held-out test partitions of datasets represented in X-AQA-40K.
For OOD evaluation, we use four datasets held out from training (T3Set, Fitness-7, KIMORE, and JIGSAWS) and include all QA samples across their released partitions.

\begin{table*}[t]
    \centering
    \caption{Task-level dataset composition. Source counts indicate the numbers of QA samples in the training and test partitions after preprocessing, with validation samples included in the test count for T3Set. X-AQA-Bench uses held-out test samples for ID evaluation and all available samples from unseen datasets for OOD evaluation. S, C, and T denote score, category, and text evaluation, respectively.}
    \label{tab:dataset}
    \small
    \setlength{\tabcolsep}{3.5pt}
    \begin{tabularx}{\textwidth}{Xllcccccc}
        \toprule
        \multirow{2}{*}{\textbf{Dataset}} & \multirow{2}{*}{\textbf{Activity}} & \multirow{2}{*}{\textbf{Task}} & \multirow{2}{*}{\textbf{Format}} & \multicolumn{2}{c}{\textbf{Source}} & \makecell{\textbf{X-AQA-40K}} & \multicolumn{2}{c}{\textbf{X-AQA-Bench}} \\
        \cmidrule(lr){5-6}\cmidrule(lr){7-7}\cmidrule(lr){8-9}
        & & & & Train & Test & Train & ID & OOD \\
        \midrule
        \multicolumn{9}{c}{\textit{Competitive Sports}} \\
        \midrule
        AQA-7 & Diving, vault, \etc & Score & S & 803 & 303 & 803 & 300 & -- \\
        MTL-AQA & Diving & Score + commentary & S,T & 1,059 & 353 & 1,059 & 339 & -- \\
        Rhythmic Gymnastics & Rhythmic gymnastics & Execution + penalty & S & 800 & 200 & 800 & 200 & -- \\
        FineFS & Figure skating & GOE grade & S & 7,469 & 2,096 & 2,000 & 300 & -- \\
        MultiSenseBadminton & Badminton & Skill & C & 1,663 & 225 & 1,663 & 6 & -- \\
        Ego-Exo4D & Basketball, soccer, \etc & Sports proficiency & C & 1,980 & 550 & 1,980 & 272 & -- \\
        T3Set & Table tennis & Coaching feedback & T & 419 & 281 & -- & -- & 700 \\
        CricketVision & Cricket batting & Technique scores & S & 6,439 & 1,755 & 2,000 & 40 & -- \\
        ExpertAF & Sports skills & Coaching feedback & T & 3,655 & 1,272 & 2,000 & 63 & -- \\
        \multirow[t]{2}{*}{CoT-AFA} & \multirow[t]{2}{*}{Martial arts} & Martial form & C & 900 & 175 & 900 & 175 & -- \\
         &  & Martial feedback & T & 699 & 175 & 699 & 175 & -- \\
        \midrule
        \multicolumn{9}{c}{\textit{Fitness Exercises}} \\
        \midrule
        Fitness-AQA & Squats, overhead presses & Technical point & C & 9,669 & 1,706 & 2,000 & 300 & -- \\
        FLAG3D & Fitness exercises & Technical point & C & 37,476 & 9,417 & 2,000 & 300 & -- \\
        \multirow[t]{2}{*}{EgoExo-Fitness} & \multirow[t]{2}{*}{Fitness exercises} & Technical point & C & 5,432 & 2,302 & 2,000 & 272 & -- \\
         &  & Overall comment & T & 636 & 272 & 636 & 272 & -- \\
        \multirow[t]{2}{*}{QEVD} & \multirow[t]{2}{*}{Fitness coaching} & Execution QA & T & 254,267 & -- & 2,000 & -- & -- \\
         &  & Coaching feedback & T & 4,408 & 1,993 & 2,000 & 300 & -- \\
        \multirow[t]{2}{*}{CoT-AFA} & \multirow[t]{2}{*}{Fitness exercises} & Fitness form & C & 1,982 & 335 & 1,982 & 300 & -- \\
         &  & Fitness feedback & T & 358 & 335 & 358 & 300 & -- \\
        Fitness-7 & Resistance exercises & Good/bad form & C & 278 & 70 & -- & -- & 348 \\
        \midrule
        \multicolumn{9}{c}{\textit{Rehabilitation}} \\
        \midrule
        KIMORE & Low-back rehabilitation & EAAQ clinical score & S & -- & 348 & -- & -- & 348 \\
        KERAAL & Low-back rehabilitation & Verdict classification & C & 540 & 491 & 540 & -- & -- \\
        REHAB24-6 & Rehabilitation exercises & Correctness & C & 727 & 345 & 727 & 23 & -- \\
        ExeCheck & Rehabilitation exercises & Correctness & C & 600 & 100 & 600 & 20 & -- \\
        LLM-FMS & Movement screening & FMS score & S & 1,451 & 361 & 1,451 & 300 & -- \\
        UpperLimb & Upper-limb rehabilitation & Completion & C & 411 & 80 & 411 & 80 & -- \\
        \midrule
        \multicolumn{9}{c}{\textit{General Skills}} \\
        \midrule
        JIGSAWS & Robotic surgery & OSATS dimensions & S & 281 & 58 & -- & -- & 339 \\
        CaptainCook4D & Cooking & Error detection & C & 2,820 & 1,397 & 2,000 & 96 & -- \\
        AIxSuture & Open suturing & OSATS global rating & S & 2,201 & 460 & 2,000 & 46 & -- \\
        CPR-Coach & Chest compression & Error detection & C & 648 & 4,118 & 648 & 300 & -- \\
        Ego-Exo4D & Cooking, music, dance & Skill proficiency & C & 3,725 & 1,035 & 2,000 & 187 & -- \\
        \multirow[t]{2}{*}{LASANA} & \multirow[t]{2}{*}{Laparoscopic training} & Skill dimensions & S & 906 & 195 & 906 & 195 & -- \\
         &  & Error detection & C & 3,521 & 763 & 2,000 & 205 & -- \\
        \midrule
        \textit{Total} & & & & & & \textbf{40,163} & \textbf{5,366} & \textbf{1,735} \\
        \bottomrule
    \end{tabularx}
\end{table*}

\subsection{Data Distributions}

\begin{figure}[t]
    \centering
    \includegraphics[width=\columnwidth]{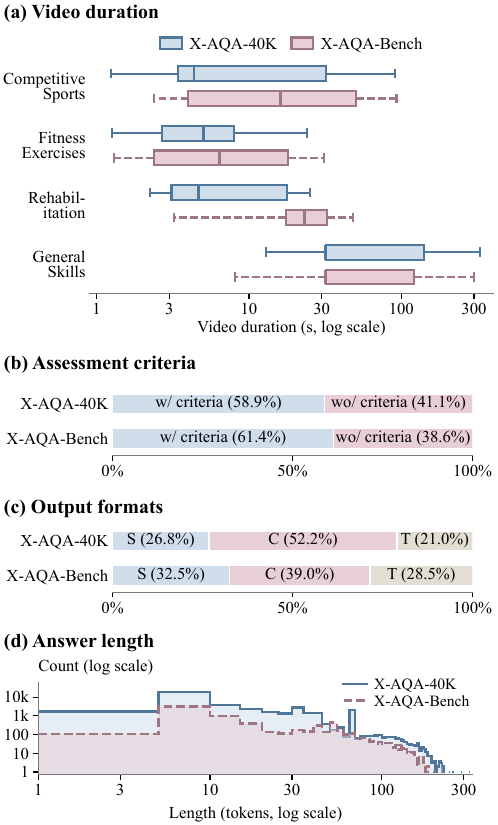}
    \caption{Data distributions of X-AQA-40K and X-AQA-Bench. (a) Video duration distributions across four domains. Video durations are reported for video samples only. (b) Percentages of samples with and without explicit assessment criteria in the prompt. (c) Shares of output components corresponding to numerical scoring (S), categorical judgment (C), and textual feedback (T). MTL-AQA samples with both a score and a comment are counted once under S and once under T. (d) Counts of complete reference answers in five-token bins. Each sample contributes one answer, tokenized with Qwen3.5-9B.}
    \label{fig:data_distributions}
\end{figure}

\Cref{fig:data_distributions} presents the distributions of video durations, assessment criteria, output formats, and reference answer lengths in X-AQA-40K and X-AQA-Bench.
Video durations vary substantially across domains.
Competitive sports and fitness exercises generally involve shorter clips, whereas general skills include longer recordings with a wider duration range.
Rehabilitation also exhibits considerable variation in clip lengths, reflecting different exercise types and recording settings.
Explicit assessment criteria are included in 58.9\% of X-AQA-40K samples and 61.4\% of X-AQA-Bench samples, providing task-specific evaluation standards across both datasets.
Categorical judgments account for the largest share of output components in X-AQA-40K (52.2\%), followed by numerical scoring (26.8\%) and textual feedback (21.0\%).
Output components in X-AQA-Bench are more evenly distributed, with 32.5\% numerical scoring, 39.0\% categorical judgment, and 28.5\% textual feedback.
Reference answer lengths also vary considerably, with short numerical or categorical responses accounting for many samples and longer textual feedback extending the distribution beyond 100 tokens.
These distributions highlight the diversity of X-AQA in temporal scale, assessment objectives, and output complexity.

\section{Experimental Details}

\subsection{Video Sampling}
\label{sec:supp_video_sampling}

We adopt an adaptive video sampling strategy to preserve temporal coverage within a maximum budget of 128 frames.
For each clip of duration $T$ seconds, we set the sampling frame rate to 16 fps for $T \leq 8$, 8 fps for $8 < T \leq 16$, and 4 fps for $16 < T \leq 32$.
For clips longer than 32 seconds, we set the sampling rate to $128/T$ fps.
The resulting frame count is constrained to the range of 4 to 128 and adjusted to an even number to accommodate the two-frame temporal patching of Qwen3.5-9B's vision encoder.
This strategy provides denser temporal sampling for short clips while maintaining coverage of longer action sequences within a fixed frame budget.

\subsection{Evaluation Protocol}

\paragraph{Prediction Parsing.}
We extract predictions according to the output schema of each task. Structured responses are parsed as JSON objects, allowing minor formatting variations such as Markdown code fences, surrounding text, and differences in field naming. Numerical predictions are validated against task-specific score ranges and increments, while categorical predictions must match the predefined labels. For multi-field outputs, all required fields must be valid. Textual responses are evaluated directly, with the comment field extracted for tasks that combine scoring and feedback. In X-AQA-Bench, only successfully parsed predictions are included in metric computation.
\Cref{tab:parsing_validity} reports the percentage of predictions that pass these checks on X-AQA-Bench.
For 14 of the 17 models, at least 95\% of predictions pass, including 100.0\% for X-AQA. VideoLLaMA3-7B has the lowest rate (76.1\%).

\begin{table}[t!]
    \centering
    \caption{Prediction parsing validity on X-AQA-Bench for the evaluated models in \cref{tab:benchmark_main}.}
    \label{tab:parsing_validity}
    \small
    \begin{tabularx}{\columnwidth}{Xr}
        \toprule
        \textbf{Model} & \textbf{Valid (\%)} \\
        \midrule
        \multicolumn{2}{c}{\textit{Open-Source Models}} \\
        Molmo2-8B & 96.5 \\
        VideoLLaMA3-7B & 76.1 \\
        InternVL3.5-8B & 99.7 \\
        Keye-VL-1.5-8B & 85.2 \\
        MiniCPM-V 4.5 & 96.1 \\
        InternVideo3-8B-Instruct & 95.9 \\
        Eagle2.5-8B & 98.8 \\
        LLaVA-OneVision-2 & 97.4 \\
        Qwen2.5-VL-7B-Instruct & 95.6 \\
        Qwen3-VL-8B-Instruct & 100.0 \\
        Qwen3-VL-8B-Thinking & 95.4 \\
        Qwen3.5-9B, Non-Thinking & 99.4 \\
        Qwen3.5-9B, Thinking & 92.6 \\
        \rowcolor{xaqamethodbg}
        \textbf{X-AQA} & 100.0 \\
        \midrule
        \multicolumn{2}{c}{\textit{Proprietary Models}} \\
        Claude Sonnet 5.5 & 100.0 \\
        Gemini 3.8 Flash & 100.0 \\
        GPT-6.1 Sol & 100.0 \\
        \bottomrule
    \end{tabularx}
\end{table}

\paragraph{X-AQA-Bench Metrics.}
We compute SRCC, macro-F1, Cohen's $\kappa$, and BERTScore-F1 as described in the main paper. For categorical and textual assessment, metrics are first computed within each data source and then averaged across sources, weighted by the number of valid predictions. Macro-F1 is averaged over classes present in the reference labels, and Cohen's $\kappa$ is unweighted. BERTScore-F1 is computed using RoBERTa-large without baseline rescaling.

\paragraph{Dataset-Specific Metrics.}
For the specialist comparisons in \cref{tab:specialists}, we evaluate X-AQA on the full test sets using the corresponding dataset-specific metrics. On MTL-AQA, the predicted execution award is converted to the final score using the annotated difficulty coefficient, following $\hat{s}=3d\hat{e}$. We report SRCC and relative $\ell_2$ error ($R$-$\ell_2$), with the latter defined as
\begin{equation}
R\text{-}\ell_2 =
\frac{1}{N}\sum_{i=1}^{N}
\left(\frac{\hat{s}_i-s_i}{s_{\max}-s_{\min}}\right)^2
\end{equation}
where $N$ is the number of valid predictions, and $s_{\max}$ and $s_{\min}$ denote the maximum and minimum reference scores in the full test set. The reported $R$-$\ell_2$ values are multiplied by 100. On Fitness-AQA, we compute error-class F1 separately for each technical point. On KIMORE, SRCC is computed independently for each of the five rehabilitation exercises. On LASANA, we report accuracy and balanced accuracy, where the latter is defined as the mean recall across classes. Both metrics are averaged equally across four representative error-detection tasks.

\section{More Qualitative Results}

\begin{figure*}[p]
    \centering
    \includegraphics[width=\textwidth]{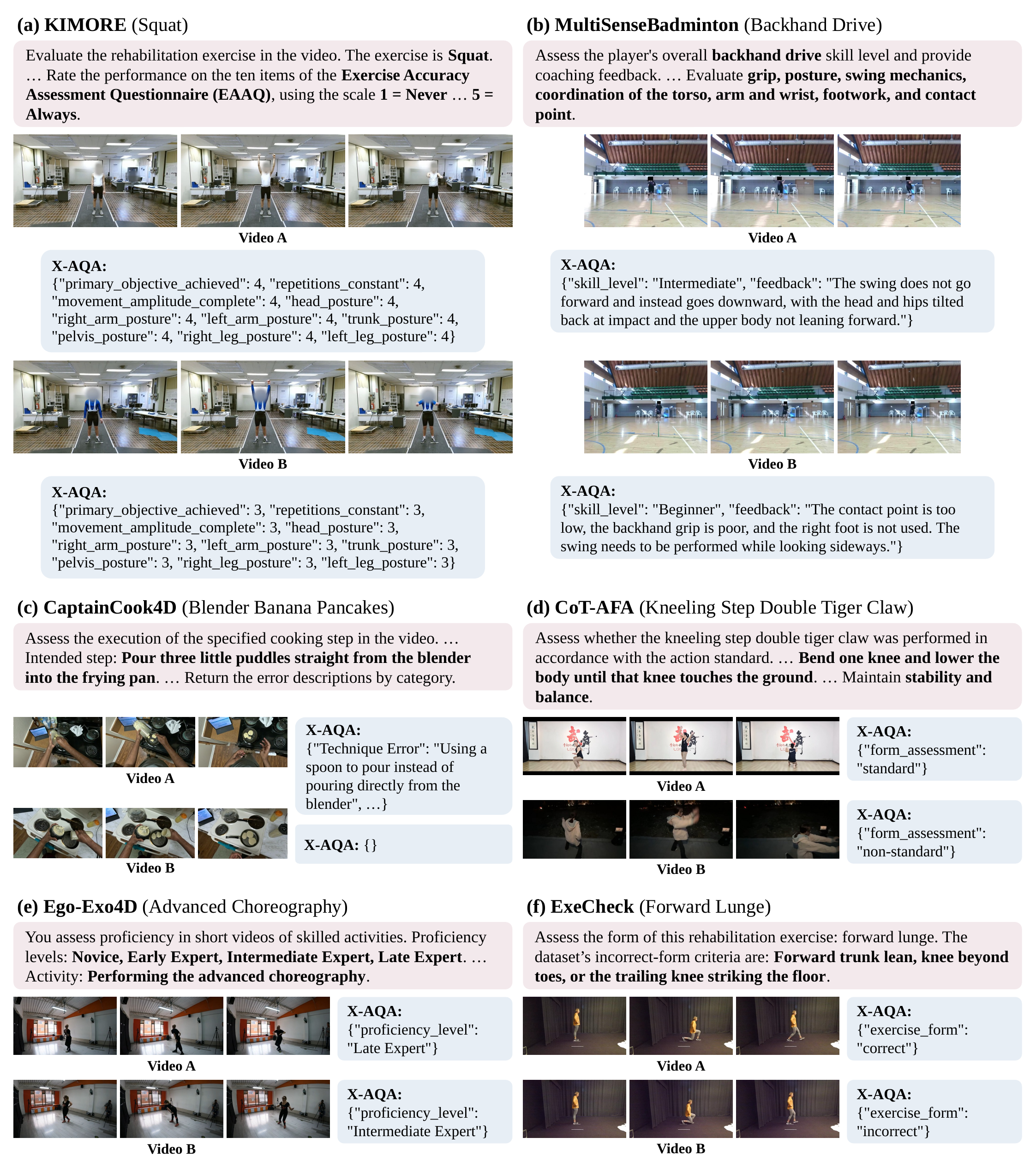}
    \caption{Additional qualitative examples of X-AQA across different videos (Part 1). Each pair of videos is evaluated using the same prompt, with X-AQA producing distinct assessments based on the observed action execution.}
    \label{fig:supp_qualitative_1}
\end{figure*}

\begin{figure*}[p]
    \centering
    \includegraphics[width=\textwidth]{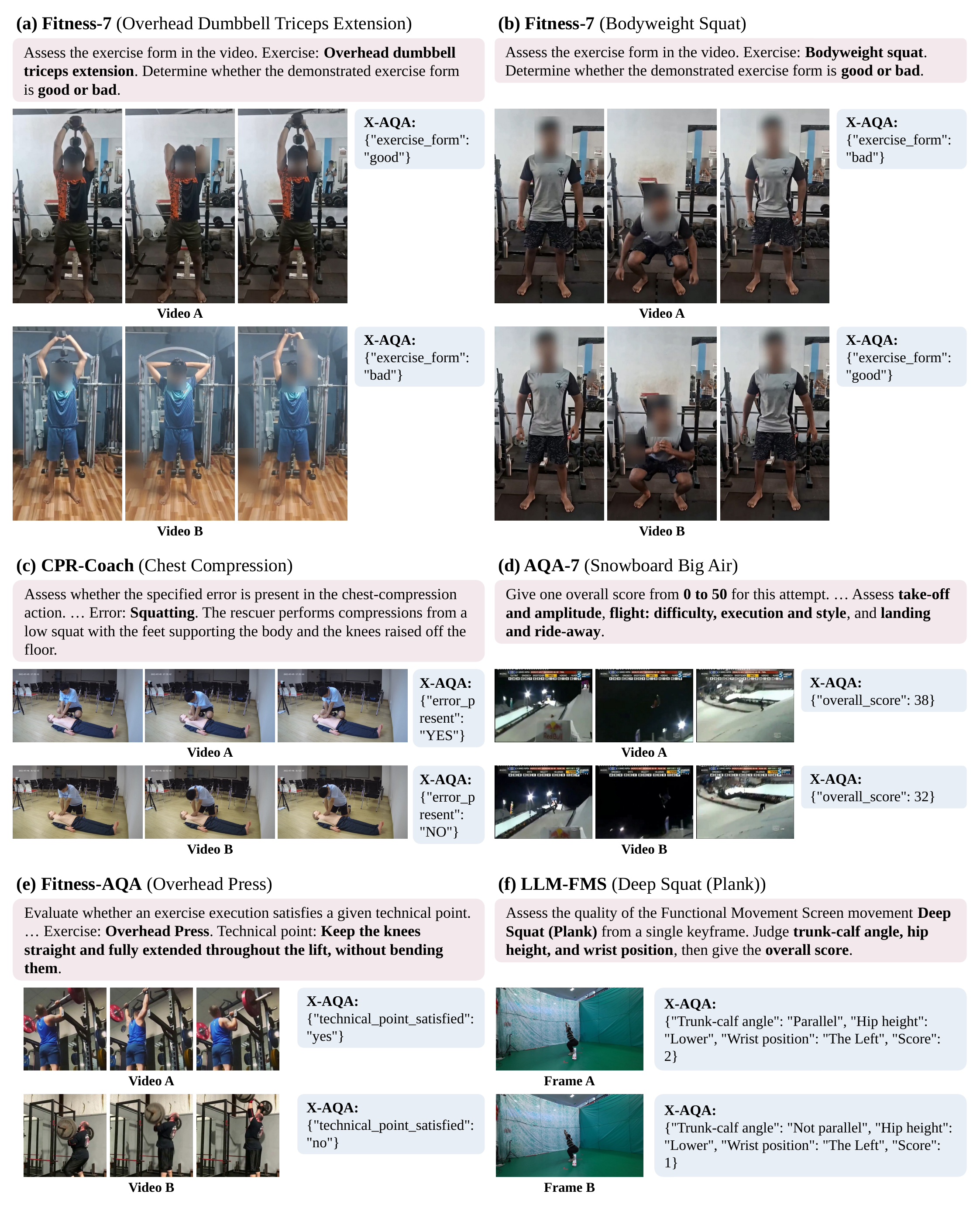}
    \caption{Additional qualitative examples of X-AQA across different visual inputs (Part 2). Each pair of visual inputs is evaluated using the same prompt, with X-AQA producing distinct assessments based on the observed action execution.}
    \label{fig:supp_qualitative_2}
\end{figure*}

\begin{figure*}[p]
    \centering
    \includegraphics[width=\textwidth]{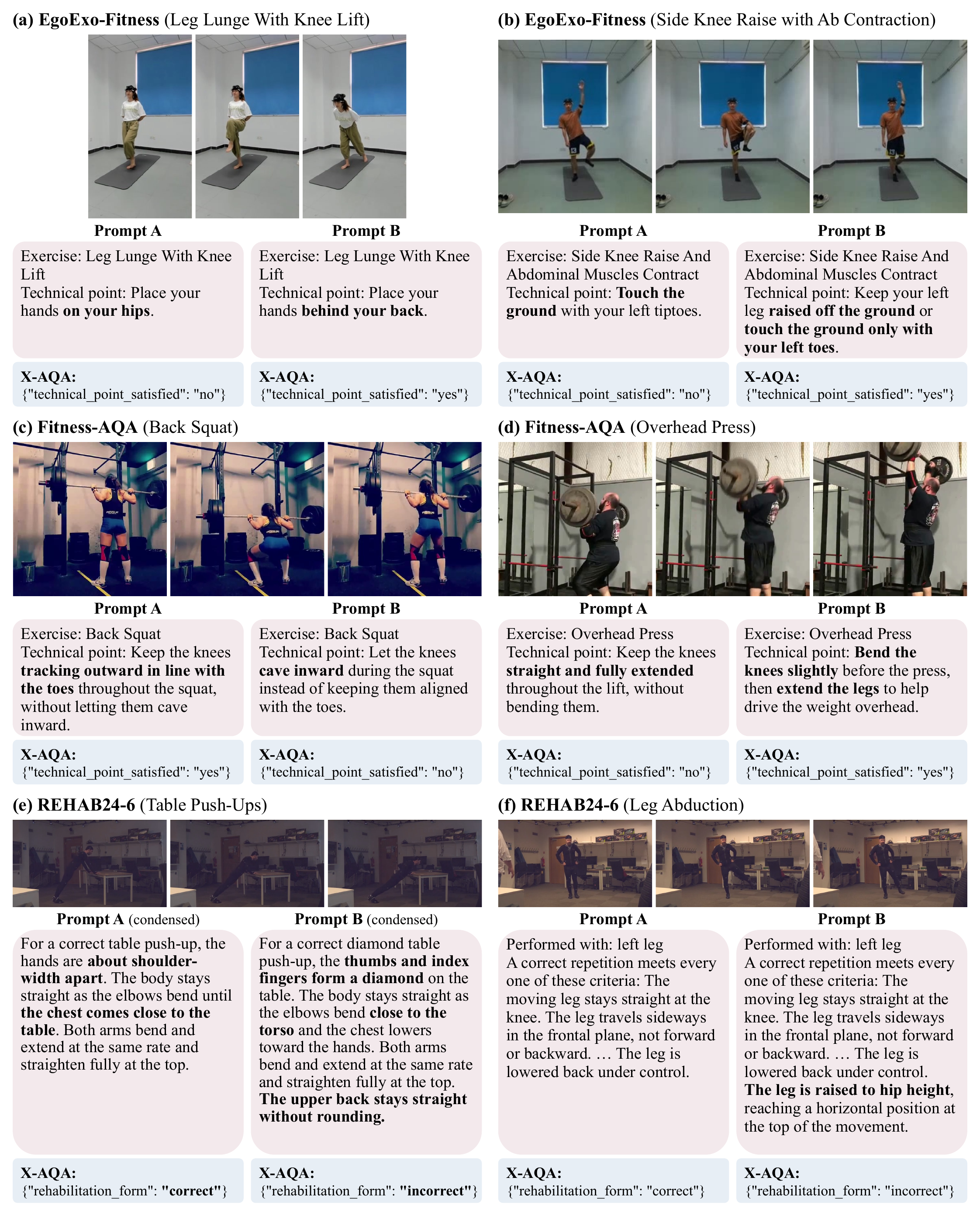}
    \caption{Additional qualitative examples of X-AQA under different assessment criteria. Each video is evaluated using two prompts with different assessment criteria, and X-AQA produces distinct judgments based on the specified requirements.}
    \label{fig:supp_qualitative_3}
\end{figure*}

\Cref{fig:supp_qualitative_1,fig:supp_qualitative_2} provide additional qualitative examples complementing those in \cref{fig:qualitative_results} in the main paper. Across diverse datasets and assessment tasks, X-AQA produces distinct assessments for different performances under the same prompt.
\Cref{fig:supp_qualitative_3} examines the complementary setting, where the same video is evaluated using different assessment criteria. X-AQA adjusts its judgments according to the specified technical requirements, correctly distinguishing whether the same action satisfies different criteria. These examples further demonstrate X-AQA's ability to assess action quality based on both observed execution and task-specific assessment criteria.

\section{Data Access and Licensing}
\label{sec:supp_data_access}

X-AQA-40K and X-AQA-Bench are constructed from existing datasets governed by their respective licenses and access agreements.
We do not redistribute the original media or annotations.
Instead, we provide detailed preprocessing code and supporting materials, including assessment criteria and dataset-specific processing configurations, to enable reconstruction of the training and evaluation data.
Users must obtain the original datasets from their respective providers and comply with the applicable licensing and access requirements.

\section{Limitations}

X-AQA establishes a unified framework for diverse AQA tasks, while several challenges remain.
First, X-AQA does not explicitly ground its predictions in visual evidence.
Linking observed movements to assessment criteria and predictions could enable more interpretable and verifiable assessments.
Second, our evaluation measures the agreement between generated feedback and reference annotations, rather than its effectiveness in improving action execution.
Extending AQA toward interactive coaching, where feedback guides and improves subsequent performance, is a promising direction for future research.

\end{document}